\documentclass[twoside,11pt]{article}

\usepackage{blindtext}

\usepackage[table,xcdraw,dvipsnames]{xcolor}

\usepackage{times}
\usepackage{epsfig}
\usepackage{amsmath}
\usepackage{amssymb}

\usepackage{svg}
\usepackage{booktabs}
\usepackage{makecell}
\usepackage{ stmaryrd }
\usepackage{subcaption}

\usepackage{jmlr2e}
\usepackage{hyperref}
\usepackage{doi}

\newcommand{\mysecref}[1]{Section~\ref{#1}}
\newcommand{\myappref}[1]{Appendix~\ref{#1}}
\newcommand{\myfigref}[1]{Figure~\ref{#1}}
\newcommand{\myeqnref}[1]{Equation~(\ref{#1})}
\newcommand{\mytabref}[1]{Table~\ref{#1}}

\newcolumntype{S}{>{\centering\let\newline\\\arraybackslash\hspace{0pt}}m{1.1cm}}
\newcolumntype{M}{>{\centering\let\newline\\\arraybackslash\hspace{0pt}}m{2cm}}
\newcolumntype{L}{>{\centering\let\newline\\\arraybackslash\hspace{0pt}}m{2.5cm}}
\newcolumntype{X}{>{\centering\let\newline\\\arraybackslash\hspace{0pt}}m{2.7cm}}

\definecolor{MyGreen}{HTML}{2ECC71}

\hypersetup{colorlinks=True, urlcolor=cyan, linkcolor=red}

\begin{document}

%%%%%%%%% TITLE
\title{A Time Series Dataset of NIR Spectra and RGB and NIR-HSI Images of the Barley Germination Process}

\author{\name Ole-Christian Galbo Engstr\o m$^{1,2,3}$ \email ocge@foss.dk \\
        \name Erik Schou Dreier$^{1}$ \email esd@foss.dk \\
        \name Birthe M\o ller Jespersen$^{4}$ \email bpmj@ucl.dk \\
        \name Kim Steenstrup Pedersen$^{2,5}$ \email kimstp@di.ku.dk \\
        \addr $^1$ FOSS Analytical A/S, Denmark\\
              $^2$ Department of Computer Science (DIKU), University of Copenhagen, Denmark\\
              $^3$ Department of Food Science (UCPH FOOD), University of Copenhagen, Denmark\\
              $^4$ University College Lilleb\ae lt (UCL), Denmark\\
              $^5$ Natural History Museum of Denmark (NHMD), University of Copenhagen, Denmark\\
}

\maketitle

\section{Introduction}\label{sec:introduction}
We provide an open-source dataset of RGB and NIR-HSI (near-infrared hyperspectral imaging) images with associated segmentation masks and NIR spectra of 2242 individual malting barley kernels. We imaged every kernel pre-exposure to moisture and every 24 hours after exposure to moisture for five consecutive days. Every barley kernel was labeled as germinated or not germinated during each image acquisition. The barley kernels were imaged with black filter paper as the background, facilitating straight-forward intensity threshold-based segmentation, e.g., by Otsu's method (\citealt{Otsu79a}). This dataset facilitates time series analysis of germination time for barley kernels using either RGB image analysis, NIR spectral analysis, NIR-HSI analysis, or a combination hereof.

This report contains an overview of the dataset in \mysecref{sec:dataset_overview}, a description of the image and spectrum acquisition in \mysecref{sec:image_acq}, and a description of the barley kernel germination process in \mysecref{sec:germination}. These sections contain sufficient information for modeling purposes for understanding the conception of the dataset. We kindly thank our sponsors in \mysecref{sec:funding}. In \myappref{app:image_processing}, we give a complete and detailed description of the image acquisition process and the methods used to identify and track individual kernels throughout several imaging sessions.

\section{Dataset overview}\label{sec:dataset_overview}
\begin{figure*}[h]
    \begin{subfigure}[t]{0.48\textwidth}
        \centering
        \includegraphics[width=\textwidth]{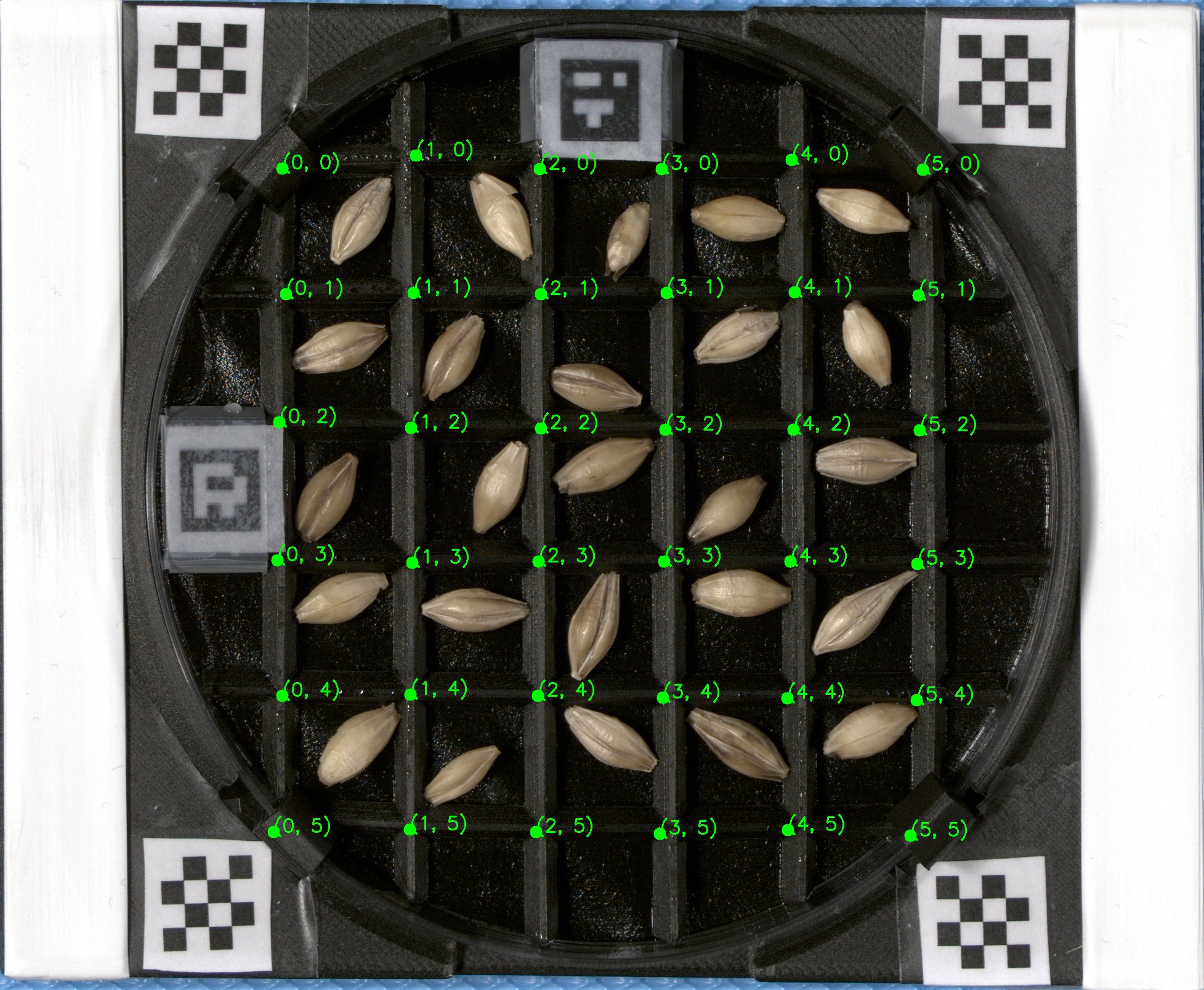}
        \caption{RGB.}
        \label{fig:dataset_rgb_grid}
    \end{subfigure}
    \hfill
    \begin{subfigure}[t]{0.48\textwidth}
        \centering
        \includegraphics[width=\textwidth]{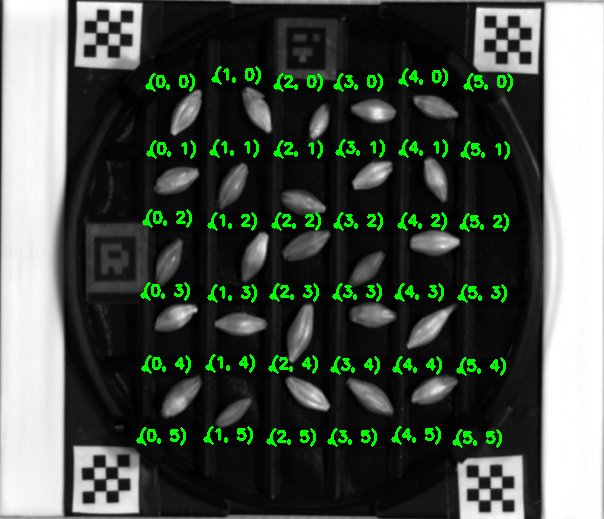}
        \caption{NIR-HSI.}
        \label{fig:dataset_hsi_grid}
    \end{subfigure}
    \caption{A Petri dish imaged at day 1. The barley kernel inside the top left square is the one shown in \mytabref{tab:grain_evolution}.}
    \label{fig:dataset_grid_img}
\end{figure*}

The dataset consists of images of 90 Petri dishes, most with 25 barley kernels. A polymer grid is placed inside each Petri dish so that each grid cell is occupied by exactly one barley kernel. During acquisition, the Petri dishes were imaged simultaneously by an RGB camera and a NIR-HSI camera, yielding RGB and NIR-HSI images of the same barley kernels under the same conditions. Each Petri dish was imaged before exposure to moisture and after exposure to moisture every 24 hours for the next five days. \mytabref{tab:grain_evolution} shows an example of tracking a single barley kernel throughout the six imaging sessions. \myfigref{fig:dataset_grid_img} shows an example of an RGB and NIR-HSI pair of images of the same Petri dish taken simultaneously.

\begin{figure*}[h]
    \begin{subfigure}[t]{0.48\textwidth}
        \centering
        \includegraphics[width=\textwidth]{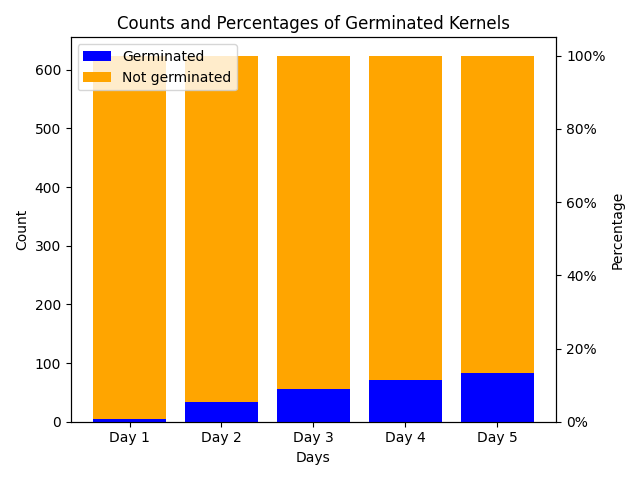}
        \caption{Prospect\_0.}
        \label{fig:Prospect_0_hist}
    \end{subfigure}
    \hfill
    \begin{subfigure}[t]{0.48\textwidth}
        \centering
        \includegraphics[width=\textwidth]{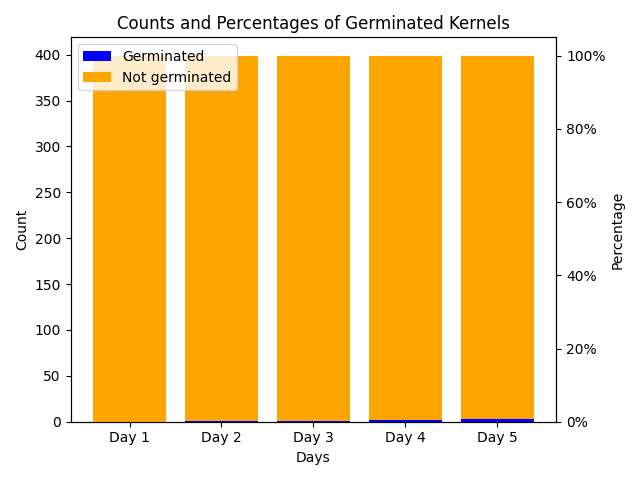}
        \caption{Prospect\_1.}
        \label{fig:Prospect_1_hist}
    \end{subfigure}
    \\
    \begin{subfigure}[t]{0.48\textwidth}
        \centering
        \includegraphics[width=\textwidth]{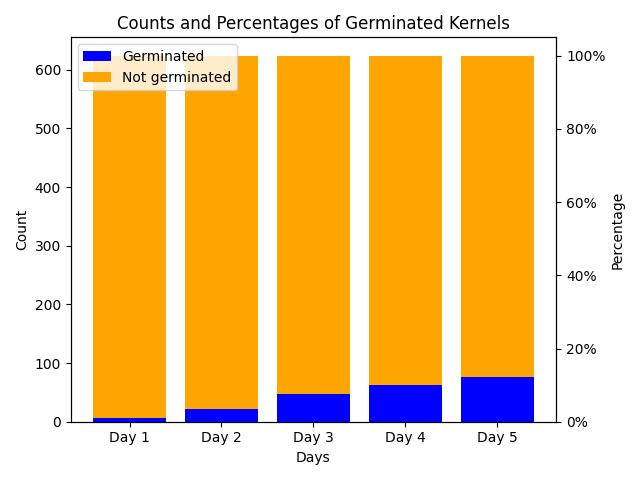}
        \caption{Laureate\_0}
        \label{fig:Laureate_0_hist}
    \end{subfigure}
    \hfill
    \begin{subfigure}[t]{0.48\textwidth}
        \centering
        \includegraphics[width=\textwidth]{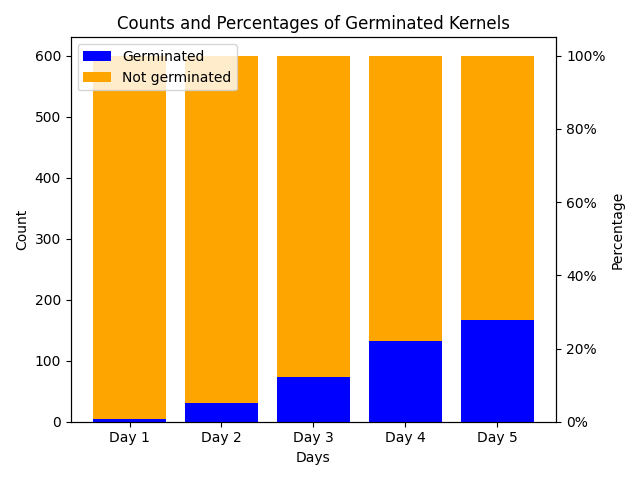}
        \caption{Laureate\_1}
        \label{fig:Laureate_1_hist}
    \end{subfigure}
    \hfill
    \begin{subfigure}[t]{0.48\textwidth}
        \centering
        \includegraphics[width=\textwidth]{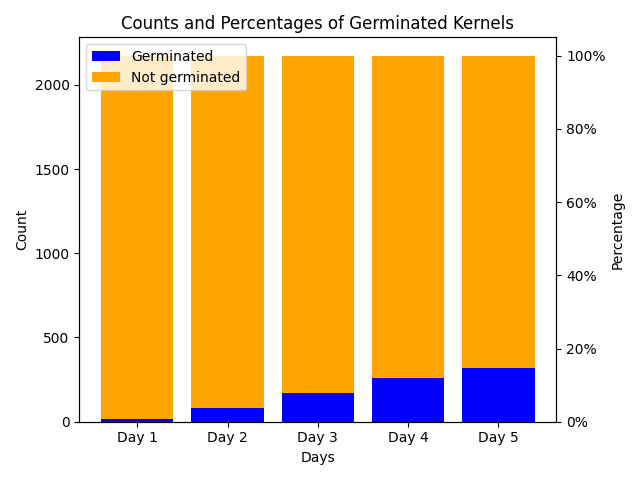}
        \caption{All varieties.}
        \label{fig:all_hist}
    \end{subfigure}
    \caption{Histograms showing how the number of germinated kernels evolves from day to day.}
    \label{fig:varieties_hists}
\end{figure*}

This dataset consists of four different samples of malting barley: two of the variety Prospect and two of the variety Laureate. \mytabref{tab:num_samples} shows the number of kernels from each variety. \mytabref{tab:germination} and \myfigref{fig:varieties_hists} show the number of kernels classified as germinated after each day.

\begin{table}[h]
\centering
\begin{tabular}{c|c|c|c|c|c|c|}
\cline{2-7}
                              & Pre moisture & Day 1 & Day 2 & Day 3 & Day 4 & Day 5 \\ \hline
\multicolumn{1}{|c|}{RGB}     & \includegraphics[width=2cm]{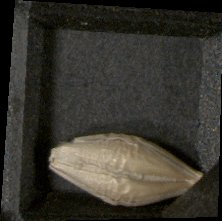} & \includegraphics[width=2cm]{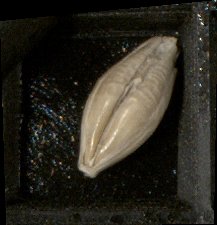} & \includegraphics[width=2cm]{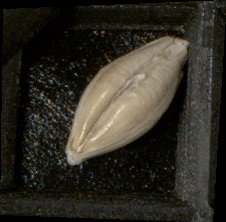} & \includegraphics[width=2cm]{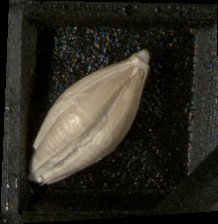} & \includegraphics[width=2cm]{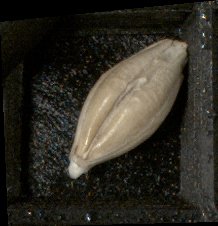} & \includegraphics[width=2cm]{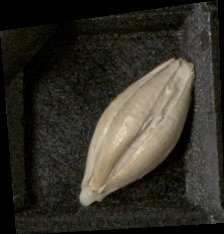} \\ \hline
\multicolumn{1}{|c|}{NIR-HSI} & \includegraphics[width=2cm]{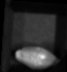} & \includegraphics[width=2cm]{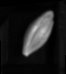} & \includegraphics[width=2cm]{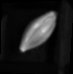} & \includegraphics[width=2cm]{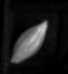} & \includegraphics[width=2cm]{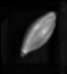} & \includegraphics[width=2cm]{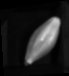} \\ \hline
\end{tabular}
\caption{A single barley kernel imaged before exposure to moisture and every 24 hours after exposure to moisture. A sprout appears on Day 2, meaning that the reference value for germination time for this specific kernel is two days.}
\label{tab:grain_evolution}
\end{table}

\begin{table}[h]
\centering
\begin{tabular}{@{}ccccc|c|@{}}
\toprule
             & \multicolumn{4}{c}{Barley Variety} \\ \cmidrule{2-6}
             & Prospect\_0  & Prospect\_1 & Laureate\_0 & Laureate\_1 & Total\\ \midrule
No. kernels  & 624 & 394 & 624 & 600 & 2242 \\
No. Petri dishes  & 25 & 16 & 25  & 24 & 90 \\ \bottomrule
\end{tabular}
\caption{Number of barley kernels and Petri dishes for each of the four barley varieties.}
\label{tab:num_samples}
\end{table}

\begin{table}[h]
\centering
\begin{tabular}{@{}cccccc|c|@{}}
\toprule
             & \multicolumn{6}{c}{Germination Day} \\ \cmidrule{2-7}
Barley Variety & 1 & 2 & 3 & 4 & 5 & Any \\ \midrule
Prospect\_0 & 4 & 29 & 23 & 15 & 12 & 83 \\
Prospect\_1 & 1 & 0 & 1 & 0 & 3 & 5 \\
Laureate\_0 & 13 & 21 & 25 & 15 & 17 & 91 \\
Laureate\_1 & 9 & 43 & 57 & 58 & 35 & 202 \\ \midrule
All  & 28 & 93 & 106 & 88 & 67 & 382 \\ \bottomrule
\end{tabular}
\caption{Number of barley kernels that germinated each day out of those shown in \mytabref{tab:num_samples}. No kernels were germinated prior to exposure to moisture.}
\label{tab:germination}
\end{table}

\subsection{Dataset availability}
We release the dataset and license it under CC BY-NC 4.0. It is available for download at \href{https://doi.org/10.17894/ucph.71c22737-8005-4588-bd74-80bf7b5ac6b4}{The University of Copenhagen's Electronic Research Data Archive}.

We provide the germination labels and images of extracted kernels, as shown in \mytabref{tab:grain_evolution}. We also provide segmentation masks for all images to distinguish the barley kernels from the background. We also provide an NIR absorbance spectrum computed as an average of the pseudo absorbance inside the segmentation mask for each NIR-HSI image. Should dataset users be interested in applying their grain kernel detection or segmentation methods, we also provide the dataset of entire Petri dish images and the coordinates of the 36 grid corners, as shown in \myfigref{fig:dataset_grid_img}.

Initially, we had envisioned 100 Petri dishes with 25 barley kernels each, totaling 2500 kernels. However, due to mold infections and human error, we lost ten Petri dishes and a few individual kernels from the remaining 90 Petri dishes. The retained dataset contains 2242 kernels. Every barley kernel in the retained dataset is imaged all six times with both cameras.

\section{Image and spectrum acquisition}\label{sec:image_acq}
\subsection{Imaging setup}
\myfigref{fig:imaging_setup} illustrates the imaging setup, which consists of a conveyor band moving a Petri dish with grain kernels underneath two line scan cameras, an RGB camera, and a NIR-HSI camera. On both ends of the setup are three 20-watt halogen light bulbs, ensuring bright illumination throughout the visual and NIR spectrum. During image acquisition, a Petri dish is placed on a plate carrying four quadratic chessboards for pixel size standardization and two windings of white PTFE (polytetrafluoroethylene) foil serving as white references. Dark references were acquired for the NIR-HSI camera by taking images with the shutter closed. For the RGB camera, no dark reference was necessary. See \myappref{app:image_processing} for further details on the image standardization process.

The RGB camera is a Basler Ace 2 Pro, while the NIR-HSI camera is a Specim FX17, with 224 uniformly distributed wavelength channels in the 900-1700nm spectrum. The RGB camera uses the standard 8 bits for each color channel, and the images are therefore stored as uint8 arrays. However, The NIR-HSI camera uses 12 bits for each wavelength channel, requiring 1.5 bytes. We store the NIR-HSI images in the mono12p format to facilitate optimal storage. This format packs two 12-bit values sequentially in three bytes (24 bits). Along with the dataset, we provide functions to convert both ways between mono12p and uint16, allowing easy analysis of uint16 images while reducing storage requirements by saving them as mono12p images. The functions are written using the Python library NumPy (\citealt{harris2020array}).

\subsection{Image acquisition and single kernel tracking}
Each Petri dish was imaged simultaneously with an RGB and an HSI camera during image acquisition. The polymer grid inside the Petri dish enables straightforward extraction of individual kernels from entire Petri dish images. To extract a single barley kernel, we rely on the polygon defined by the coordinates of the four surrounding grid points. To extract a rectangular image, we compute the circumscribed rectangle of the polygon, making sure to mask out (set to zero) any part of the rectangle outside the polygon. This process yields single kernel images, as shown in \mytabref{tab:grain_evolution}.

The ArUco codes are fixed on the grids, which are, in turn, fixed on the Petri dishes. Thus, based on the location of the ArUco codes' eight corners (four per code), we can use random sample consensus (RANSAC, \citealt{fischler1981ransac}) to estimate an affine transformation matrix from a previous image of the same Petri dish to the current image. Based on the affine transformation matrix, we can track each grid cell and each barley kernel across multiple imaging sessions and between the RGB and NIR-HSI images. \myappref{app:image_processing} contains a more detailed description of the single kernel tracking and grid detection.

\subsection{Segmentation mask and spectrum acquisition}
For each barley kernel, RGB and HSI alike, we compute a segmentation mask by applying Otsu's method (\citealt{Otsu79a}) on a grayscale version of the image, computed as a uniform average over all the spectral channels. After applying Otsu's method, we compute all connected components of the foreground and retain only the largest set of connected components. Chaining Otsu's method with connected components attempts to remove any reflections in the moist filter paper that might otherwise be classified as foreground by Otsu's method.

We use the segmentation mask on the NIR-HSI images to compute a mean pseudo-absorbance spectrum for each kernel image. This computation takes the negative logarithm of every pixel inside the mask in the original reflectance image and then averages the result. Sometimes, the hyperspectral camera measured zero reflectance in one or more wavelength channels of a pixel. We deemed such zero measurements to be sensor faults. Since the logarithm of zero is undefined, we instead infer the reflectance value by linear interpolation between neighboring wavelength channels in the same pixel. This procedure alleviated all cases of measured zero reflectance in this dataset. Additionally, as the hyperspectral camera has reduced sensitivity in the first and last then spectral channels, we remove these from the mean spectra to facilitate easier downstream analysis for chemometricians.

\begin{figure*}
    \centering
    \begin{subfigure}[b]{\textwidth}
        \centering
        \includegraphics[width=\textwidth]{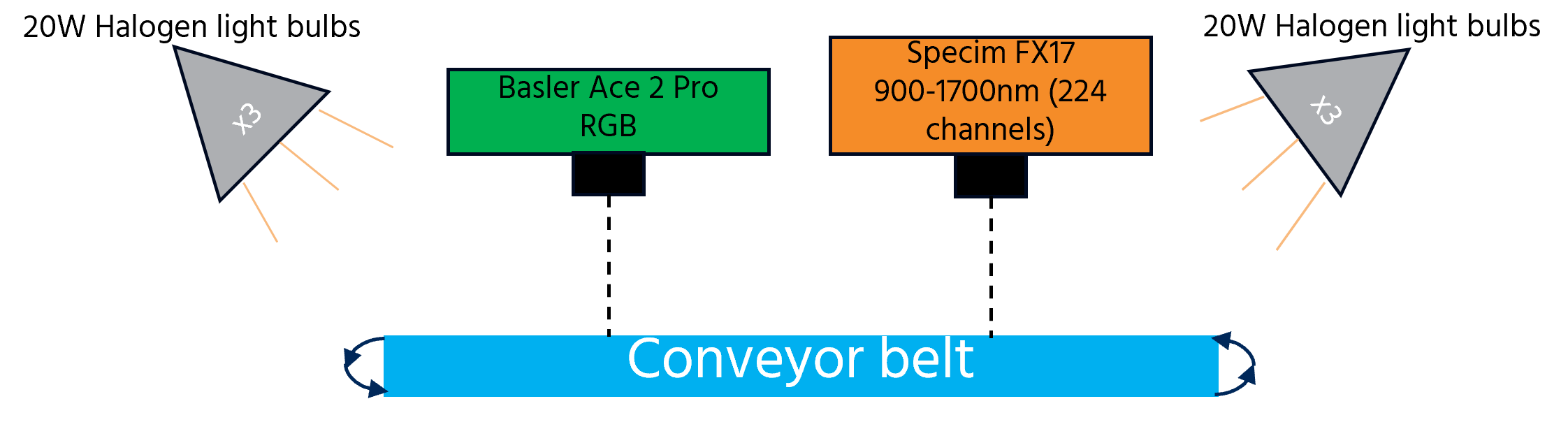}
        \caption{An illustration of the imaging setup. The plate carrying the Petri dishes, the white reference, and the chessboards moves along the conveyor belt at a constant velocity. The conveyor belt is illuminated from each side by three 20W halogen light bulbs. The RGB and HSI cameras both operate in line-scan mode.}
    \end{subfigure}
    
    \vspace{1em}

    \begin{subfigure}[t]{0.63\textwidth}
        \centering
        \includegraphics[width=\textwidth]{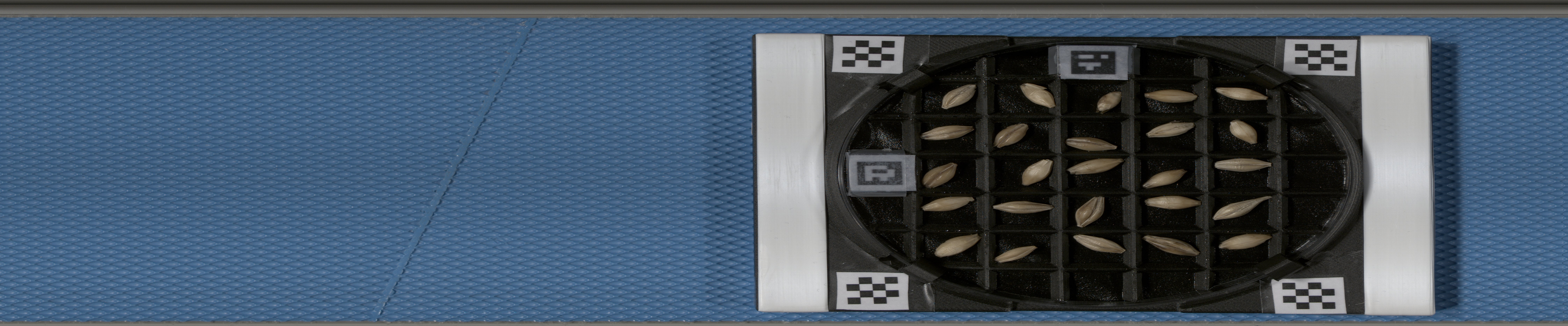}
        \caption{An example raw output from the RGB camera.}
    \end{subfigure}
    \begin{subfigure}[t]{0.33\textwidth}
        \centering
        \includegraphics[width=\textwidth]{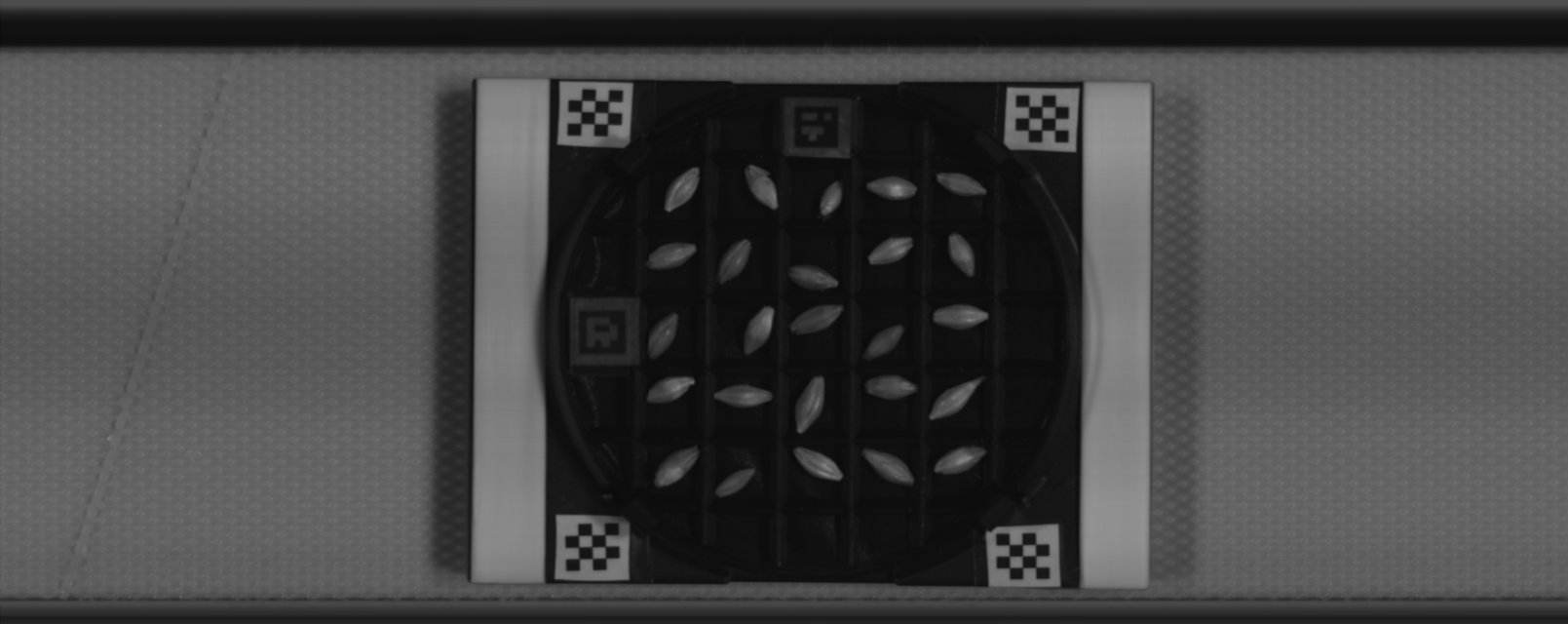}
        \caption{An example raw output from the HSI camera, converted to grayscale for visualization purposes.}
    \end{subfigure}
    \caption{Illustration of the imaging setup and example raw camera outputs.}
    \label{fig:imaging_setup}
\end{figure*}

\begin{figure*}[htbp]
    \centering
    % First row
    \begin{subfigure}[b]{0.48\textwidth}
        \centering
        \includegraphics[width=\textwidth]{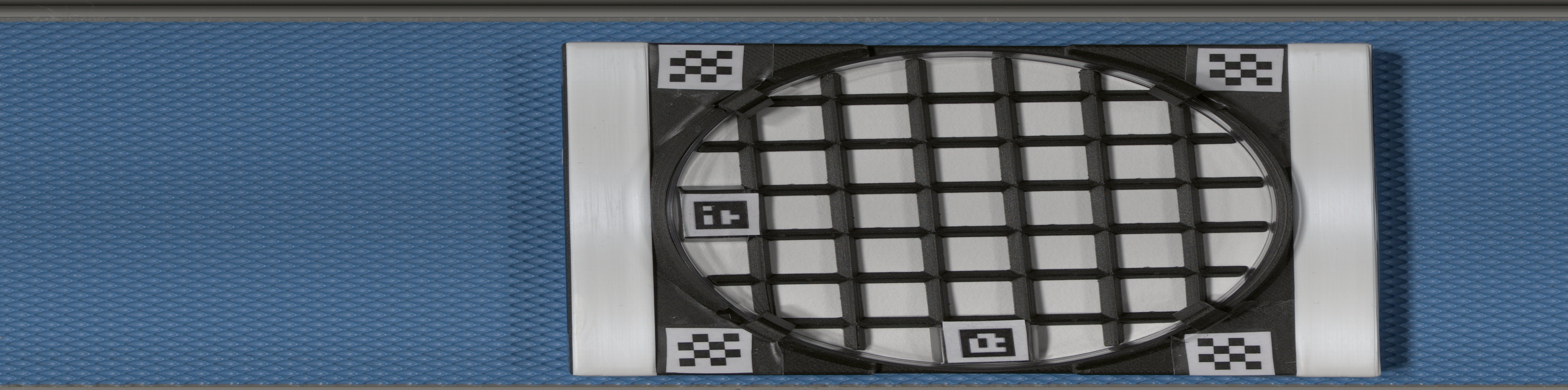}
        \caption{Raw RGB image with white filter paper for easier localization of black grid lines.}
        \label{fig:raw_rgb_reference}
    \end{subfigure}
    \hfill
    \begin{subfigure}[b]{0.48\textwidth}
        \centering
        \includegraphics[width=\textwidth]{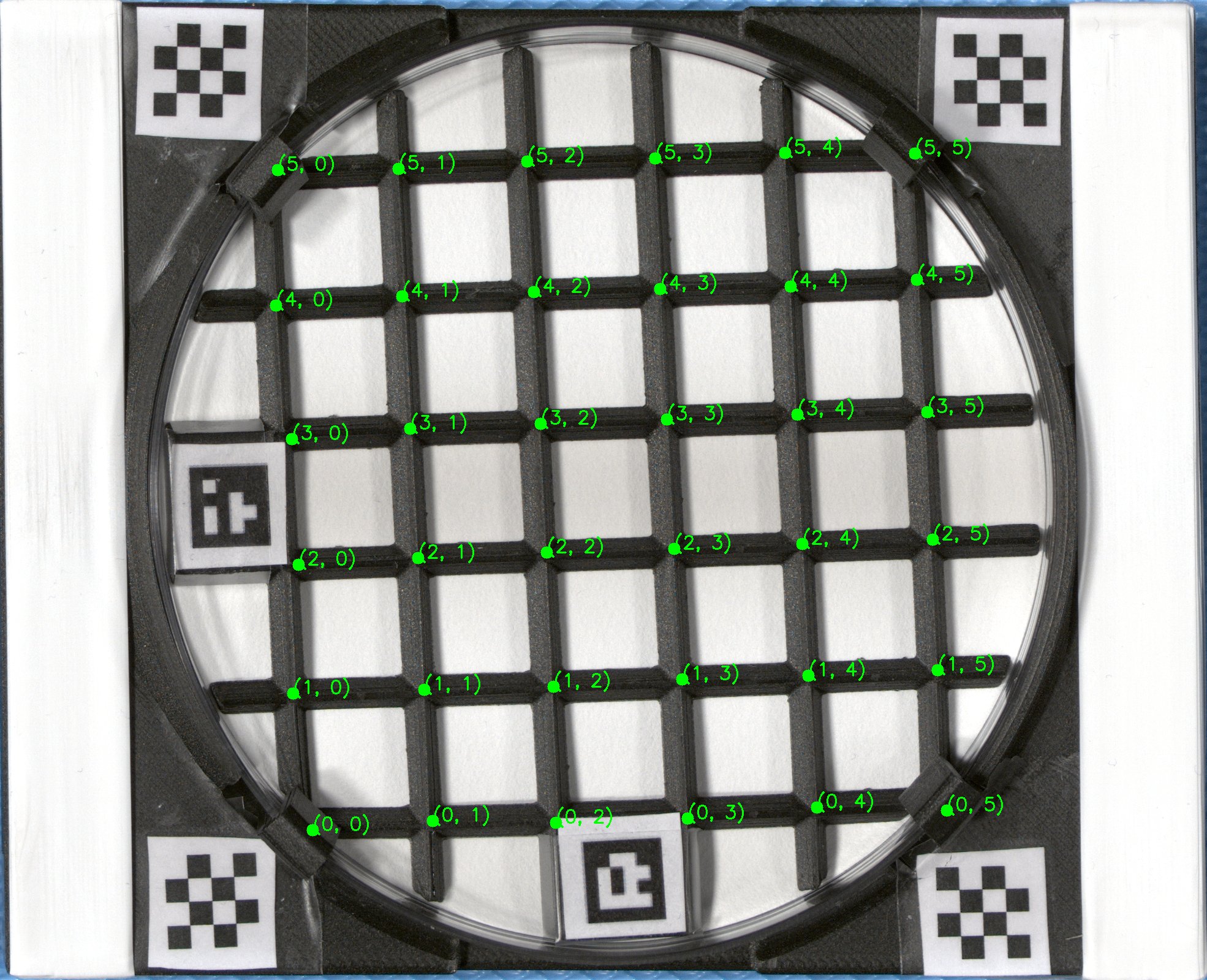}
        \caption{\myfigref{fig:raw_rgb_reference} after extracting the plate, applying white balance correction, size correction, and the grid finding algorithm.}
        \label{fig:reference_grid_rgb}
    \end{subfigure}
    
    \vspace{1em}
    
    % Second row
    \begin{subfigure}[t]{0.48\textwidth}
        \centering
        \includegraphics[width=\textwidth]{figures/pipeline/dish_26_rgb_day_1/grid_img.jpg}
        \caption{The Petri dish from \myfigref{fig:raw_rgb_reference} with black instead of white filter paper for higher contrast with the barley kernels. An affine transformation matrix from \myfigref{fig:reference_grid_rgb} to this image is computed based on the 8 corners of the ArUco codes. This affine transformation matrix is then used to move the grid from \myfigref{fig:reference_grid_rgb} to this image. This image is acquired 1 day after exposure to moisture.}
        \label{fig:rgb_grid_day_1}
    \end{subfigure}
    \hfill
    \begin{subfigure}[t]{0.48\textwidth}
        \centering
        \includegraphics[width=\textwidth]{figures/pipeline/dish_26_hsi_day_1/grid_img.jpg}
        \caption{A grayscale visualization of the HSI image corresponding to the RGB image in \myfigref{fig:rgb_grid_day_1}. The placement of the grid is derived based on an affine transformation between the centers of the 4 chessboards in \myfigref{fig:rgb_grid_day_1} and those in this image.}
        \label{fig:hsi_grid_day_1}
    \end{subfigure}
\end{figure*}

\begin{figure*}[htbp]\ContinuedFloat
    % Last row with smaller images aligned to the center of their respective columns
    \begin{subfigure}[b]{0.48\textwidth}  % First image aligned with the left column
        \centering
        \includegraphics[width=0.1\textwidth]{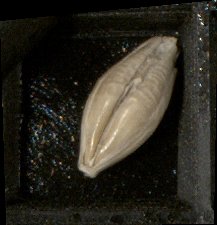}  % Smaller image
        \caption{A cutout of the grid cell in \myfigref{fig:rgb_grid_day_1} defined by a polygon between the grid points $\{(0,0), (1,0), (1,1), (0,1)\}$.}
    \end{subfigure}
    \hfill
    \begin{subfigure}[b]{0.48\textwidth}  % Second image aligned with the right column
        \centering
        \includegraphics[width=0.1\textwidth]{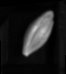}  % Smaller image
        \caption{A cutout of the grid cell in \myfigref{fig:hsi_grid_day_1} defined by a polygon between the grid points $\{(0,0), (1,0), (1,1), (0,1)\}$.}
    \end{subfigure}
    
    \caption{An overview of the image processing pipeline using Petri dish 26 as an example. Notice how the grid corners are uniquely defined by the detection in the RGB image in \myfigref{fig:reference_grid_rgb}. Localization of the same grid in a subsequent RGB or HSI image can be derived based on corresponding affine transformation matrices based on the ArUco codes' corners and the chessboards' centers, respectively.}
    \label{fig:pipeline_summary}
\end{figure*}

\section{Germination}\label{sec:germination}
To provoke barley germination, we used The European Brewery Convention (EBC) method 3.6.2 (\citealt{ebc}). The following is a description of the method. Two layers of black filter paper were placed at the bottom of each Petri dish, atop which we placed the grid and barley kernels. After the initial image acquisition (pre-moisture), we distributed 4ml tap water evenly across the filter paper, thus allowing the barley kernels to absorb the moisture. Between imaging sessions (Day 1 to Day 5), each Petri dish was contained inside a zip-lock plastic bag to contain the moisture and diminish vaporization. The plastic bags were carefully placed inside a box in a dark room at room temperature. After image acquisition, a human expert labeled each barley kernel as germinated or not germinated based on the RGB images.

Readers and users familiar with barley germination may wonder why so few kernels germinate (visualized in \myfigref{fig:varieties_hists}). After collecting the dataset, we investigated, revealing the reason. Underneath the 3D printed polymer grid used in Petri dishes (shown in \myfigref{fig:dataset_grid_img}) were glue remains from a glue stick used in the 3D printing process. The glue severely inhibits the barley kernels' ability to germinate.

We investigated if there was a difference in germination degree between black versus white filter paper, imaging (placing the Petri dishes under the bright illumination from the halogens shown in \myfigref{fig:imaging_setup}) versus no imaging and using a grid versus not using a grid. This investigation showed a significant difference between using and not using the grid. We subsequently tried to isolate whether it was due to the grid's material or the underlying glue remains. The investigation clearly showed that the glue remains were the inhibiting factor.

Still, even with the glue inhibiting the germination process, we envision that the dataset can still be used for modeling the germination process. In particular, the dataset can be used directly to classify whether or not a barley kernel is germinated. However, when training models to predict the evolution of the germination process, the bias of this dataset should be considered if transferring to applications outside this dataset. Predicting whether a given barley kernel will germinate in the future depends on the subsequent treatment that the barley kernel receives. In this dataset, the barley kernel received treatment with water contaminated with glue remains, which is not representative of the typical process without glue remains. Thus, a given barley kernel may have germinated given the proper treatment, although it would not have germinated with the treatment we gave.

\section{Funding}\label{sec:funding}
This dataset has been acquired as part of an industrial Ph.D. project receiving funding from FOSS Analytical A/S and The Innovation Fund Denmark. Grant number 1044-00108B.

\section{Acknowledgments}\label{sec:acknowledgments}
We would like to thank our amazing colleagues at FOSS Analytical A/S for their help in constructing the imaging setup and writing its control software, designing and constructing the plate carrying the Petri dishes and grids used to separate individual kernels, and acquiring the images.

\bibliography{bibliography}

\appendix
\section{Image processing}\label{app:image_processing}
In this section, we describe how to identify individual grid cells and track them across different images of the same Petri dish. We use Petri dish 26 as an example, but the process is the same for every Petri dish.

The overall strategy is to initially detect each of the unique grids using the RGB camera once. Then, for subsequent RGB images, we localize the grid based on an affine transformation from the initially detected grid and the current RGB image. This affine transformation is based on the eight corners of the ArUco codes. Any affine transformation of the grid can be computed based on the ArUco codes as they are fixed on the grid.

We can not count on the detection of the ArUco codes to locate the grid on an HSI image. This is because the paper on which the ArUco codes are printed absorbs water and thus becomes very dark in the NIR wavelengths, diminishing the contrast between the otherwise white paper and the black ArUco code. Instead, we opt for an affine transformation based on detecting the plate-fixed chessboards on the HSI image and its corresponding RGB image - the latter for which we derived the grid coordinates described above. As the Petri dish does not move relative to the plate between acquiring a pair of RGB and HSI images, we can use this affine transformation matrix to localize the grid coordinates on an HSI image, given the grid coordinates on its corresponding RGB image.

Using affine transformations instead of executing the entire grid-detection pipeline for every image acquisition gives robustness and consistency in localizing grid coordinates across multiple images of the same physical Petri dish. Additionally, this strategy allows switching between filter papers (backgrounds) to maximize contrast between the foreground. Therefore, we use a white filter paper for the initial grid detection to maximize contrast with the black grid. Then, we use black filter paper for subsequent images to maximize contrast with the brown barley kernels and, importantly, the white sprouts.

In \mysecref{sec:initial_grid_detection}, we show how to initially detect the grid coordinates from an RGB image with a white filter paper background. Then, in \mysecref{sec:rgb_grid_localization} and \mysecref{sec:hsi_grid_localization}, we show how to localize the same grid in subsequent RGB and HSI images, respectively, with grain and black filter paper background. This strategy allows us to track individual grid cells (and therefore barley kernels) across different images of the same Petri dish.

\subsection{Initial grid detection}\label{sec:initial_grid_detection}
The entire process of grid detection is outlined in \myfigref{fig:reference_pipeline}. The raw RGB camera output (\myfigref{fig:reference_raw_rgb}) is grayscaled (\myfigref{fig:reference_raw_gray}) to allow for an Otsu thresholding (\citealt{Otsu79a}) so we can identify the two largest connected components (\myfigref{fig:reference_raw_largest_components}), corresponding to the location of the two white references (\myfigref{fig:reference_white_reference}). For each color channel, we can then compute an average intensity for each row of the white reference (\myfigref{fig:reference_white_reference_plot}). As the camera is used in line-scan mode, we can use this result to estimate how much light is hitting each point on the plate. We use the third quantile of the intensity of each row to get a robust value against shadows and small dust particles that may have stuck to the white references. We also use the location of the white references as a bounding box to extract the plate (\myfigref{fig:reference_plate}). We can then use the white-reference intensities to adjust the intensity for each row to mitigate the effects of non-uniform illumination (\myfigref{fig:reference_plate_color_corrected}). The intensity correction is based on \myeqnref{eq:white_dark_correction}. Here $i,j,c$ are the row (height), column (width), and channel (depth) indices, respectively. $W$ and $D$ are white and dark references, respectively, and both are 2-dimensional containing, representing the measured white and dark intensities for each channel, $c$, at each row, $i$. For the RGB images $D_{i,c}=0$ for all $i,c$ as the RGB camera has no dark current.\footnote{We blocked the RGB camera's lens with a black cap and let it capture several million pixels while having the halogen illumination system active. Every single pixel in the resulting image had a value of $\{$R,G,B$\}$ = $\{0,0,0\}$.}

\begin{equation}\label{eq:white_dark_correction}
    I^{\text{corrected}}_{i,j,c} = \frac{I^{\text{raw}}_{i,j,c} - D_{i,c}}{W_{i,c} - D_{i,c}}
\end{equation}

We can then detect the chessboard squares, and because they are squares in the physical world, we can use their sizes to estimate the factor by which to squish the horizontal dimension of the image. Because the horizontal dimension is the largest, and the camera works in line-scan by acquiring one vertical line at a time, there is no loss of information. We downsize the image using bilinear interpolation (\myfigref{fig:reference_plate_size_and_color_corrected}). The process described thus far finds the plate, applies the white standardization, and size correction. This provides a strong starting point for the subsequent analysis to identify the grid cells.

First, we detect the two ArUco codes. Each is identified by its unique id and the four corners. These will be used for subsequent analysis. Then, we perform smoothing followed by edge detection, the result of which is the input to a Hough Circle Transform (\citealt{duda1972use}). After thresholding the circle size, if multiple circles remain, we choose the one with its center closest to the center of the image (\myfigref{fig:reference_hough_circle}). The goal is to segment the Petri dish for further analysis (\myfigref{fig:reference_circle_mask}). After segmenting the Petri dish, we perform smoothing and subsequent edge detection, then masking the ArUco codes. We feed the result to a Hough Line Transform (\citealt{duda1972use}) to detect the borders of the grid lines \myfigref{fig:reference_hough_lines}. We average the detected borders to get an estimate of the location of the grid lines (\myfigref{fig:reference_avg_hough_lines}). We then compute the intersection between all the grid lines, retaining only intersections inside the image's borders (\myfigref{fig:reference_intersections}).

Now, we have detected points that correspond to grid intersections. Further analysis will refine this estimate to find a unique orientation of the grid. Notice how points in the far left of \myfigref{fig:reference_intersections} do not correspond to grid intersections. However, the subsequent analysis is robust enough to consider this.

Consider now \myfigref{fig:reference_vecs_img}. For each ArUco code, we compute its center by averaging the coordinates of its corners. Then, from the center, we compute a vector (red) to the center of the Hough Circle found in \myfigref{fig:reference_hough_circle}. We then compute vectors from the ArUco codes center to all (green) intersections. We retain the closest two that form an angle of less than 90 degrees with the corresponding (red) vector to the center.

For each ArUco code, consider now the pair of points found before. These two points each lie on precisely two lines, and the two points share exactly one of the lines. We now choose an arbitrary point from the pair (the final grid is independent of the point choice). Now, choose the line 
\emph{not shared by the other point}. In \myfigref{fig:reference_vecs_img_with_lines}, we paint magenta (yellow) the line chosen from the point pair of ArUco code with the lowest (highest) id.

Next, consider \myfigref{fig:vecs_img_with_lines_and_intersections}. On the magenta (yellow) line, we find all (green) intersection points that lie on this line. We then use the previously computed vectors from the center of the ArUco code with the lowest (highest) id to these points. We retain those that form an angle of less than 90 degrees with the (red) vector from the ArUco code to the center. We sort the vectors by length and retain only the six closest vectors, shown in yellow (magenta).

Now, for each point on the yellow (magenta) vectors, identify the other line on which it lies. That is, not the magenta (yellow) line already detected. This other line is then marked yellow (magenta), and it is the line for which the y-coordinate (x-coordinate) of the grid is constant. The y-coordinate (x-coordinate) value is determined by the length of the vector associated with the point used to detect it. This is all visualized in \myfigref{fig:reference_grid_lines_img}.

Now, we can compute the intersection points between all magenta and yellow lines. Each point can then be given a unique $(x,y)$ coordinate based on the magenta and yellow line on which it lies. To robustify the grid detection, we move the points to a local minimum in intensity space as a final step. This step adjusts the grid points to lie in the middle of a physical grid intersection. We do not adjust points with coordinates $(x,y) \in \{(0,0), (0,5), (5,0), (5,5), (3,0), (2,0), (0,3), (0,2)\}$. This is because these points lie near the ArUco codes or the borders of the Petri dish, and, as such, a local intensity minimum is not a good estimate of the physical grid intersection. The final grid detection is shown in \myfigref{fig:reference_grid_img}.

\begin{figure*}[htbp]
    \centering
    % First row
    \begin{subfigure}[b]{\textwidth}
        \centering
        \includegraphics[width=\textwidth]{figures/pipeline/reference_img_26/raw_bgr_img.jpg}
        \caption{The raw image output by the RGB camera with white filter paper for easier localization of black grid lines.}
        \label{fig:reference_raw_rgb}
    \end{subfigure}
    
    \vspace{1em}
    
    % Second row
    \begin{subfigure}[t]{\textwidth}
        \centering
        \includegraphics[width=\textwidth]{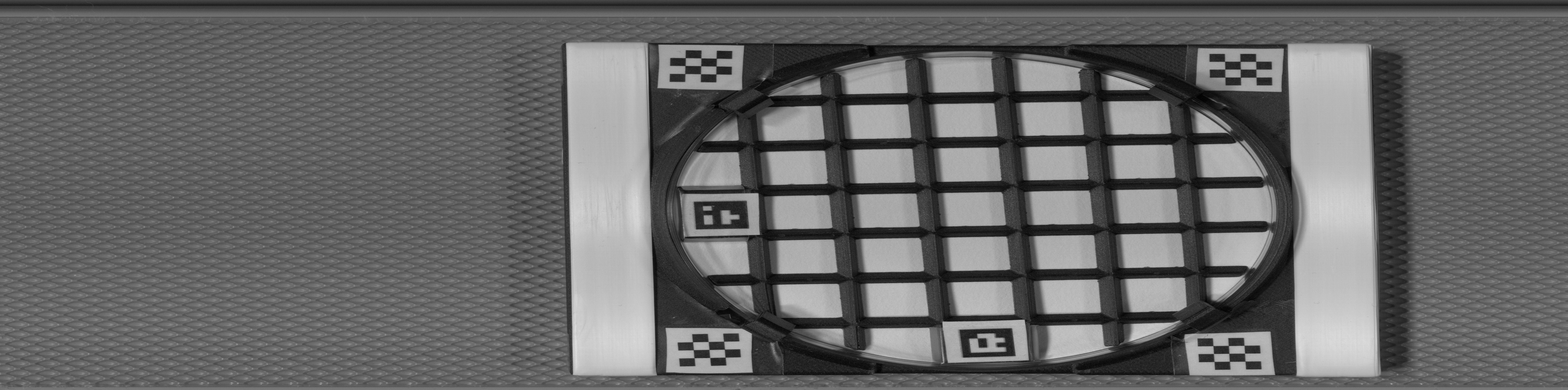}
        \caption{Grayscale version of \myfigref{fig:reference_raw_rgb} achieved by averaging the color channels.}
        \label{fig:reference_raw_gray}
    \end{subfigure}

    \vspace{1em}

    % Third row
    \begin{subfigure}[t]{\textwidth}
        \centering
        \includegraphics[width=\textwidth]{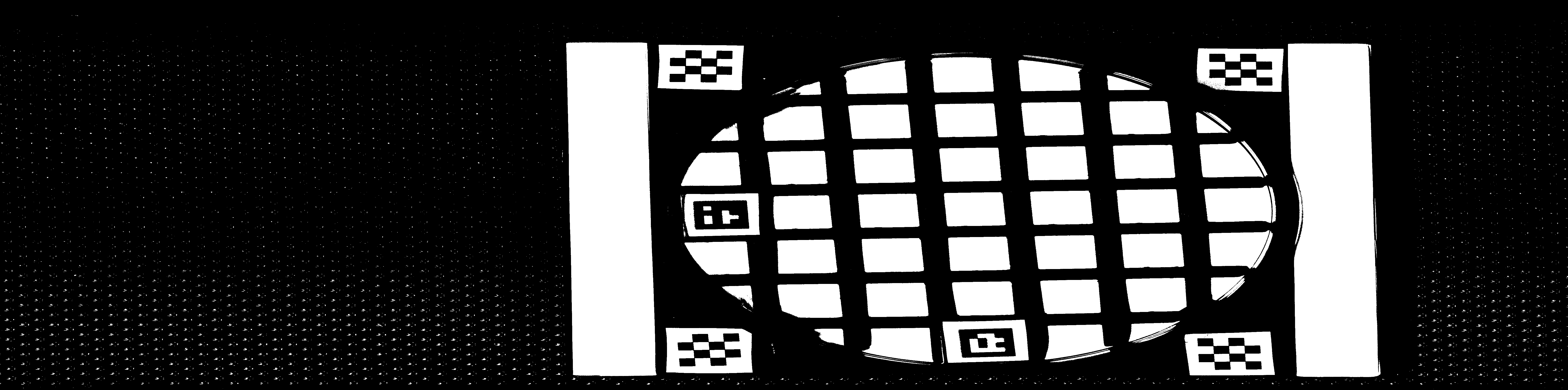}
        \caption{Binary version of \myfigref{fig:reference_raw_gray} computed using Otsu's method (\citealt{Otsu79a}).}
        \label{fig:reference_raw_binary}
    \end{subfigure}
\end{figure*}

\begin{figure*}[htbp]\ContinuedFloat
    \centering
    \begin{subfigure}[t]{\textwidth}
        \centering
        \includegraphics[width=\textwidth]{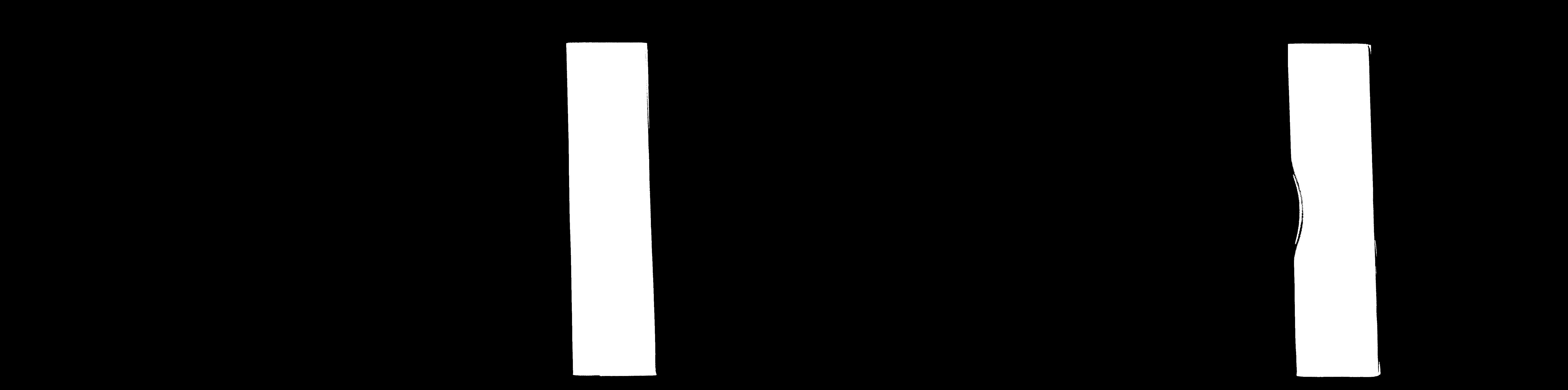}
        \caption{The two largest connected components from \myfigref{fig:reference_raw_binary}. They correspond to the location of the two white references.}
        \label{fig:reference_raw_largest_components}
    \end{subfigure}

    \vspace{1em}
    
    \begin{subfigure}[b]{\textwidth}
        \centering
        \includegraphics[width=\textwidth]{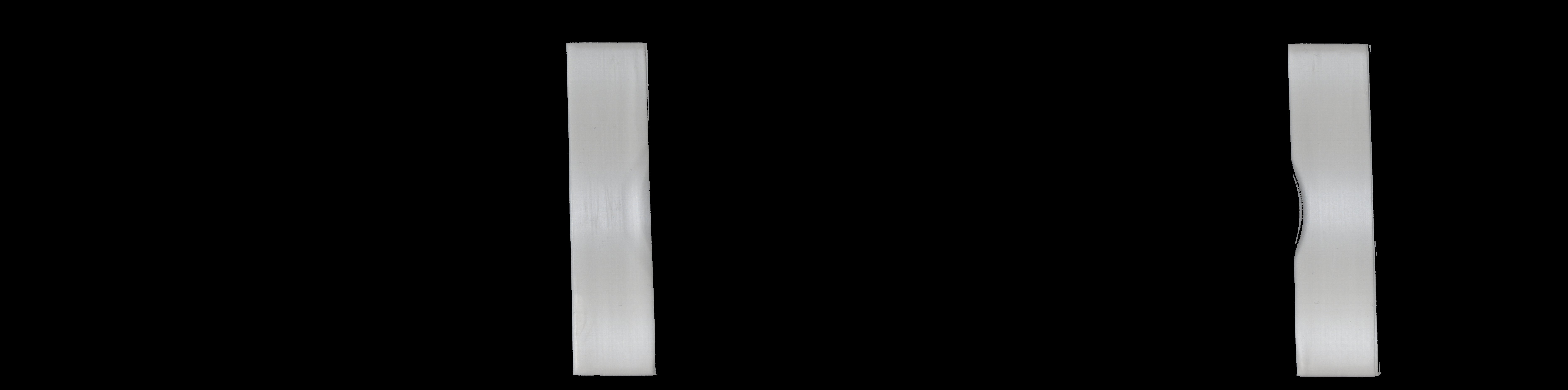}
        \caption{Applying \myfigref{fig:reference_raw_largest_components} as a binary mask to \myfigref{fig:reference_raw_rgb} to extract the white references.}
        \label{fig:reference_white_reference}
    \end{subfigure}
    
    \vspace{1em}
    
    \begin{subfigure}[b]{\textwidth}
        \centering
        \includegraphics[width=\textwidth]{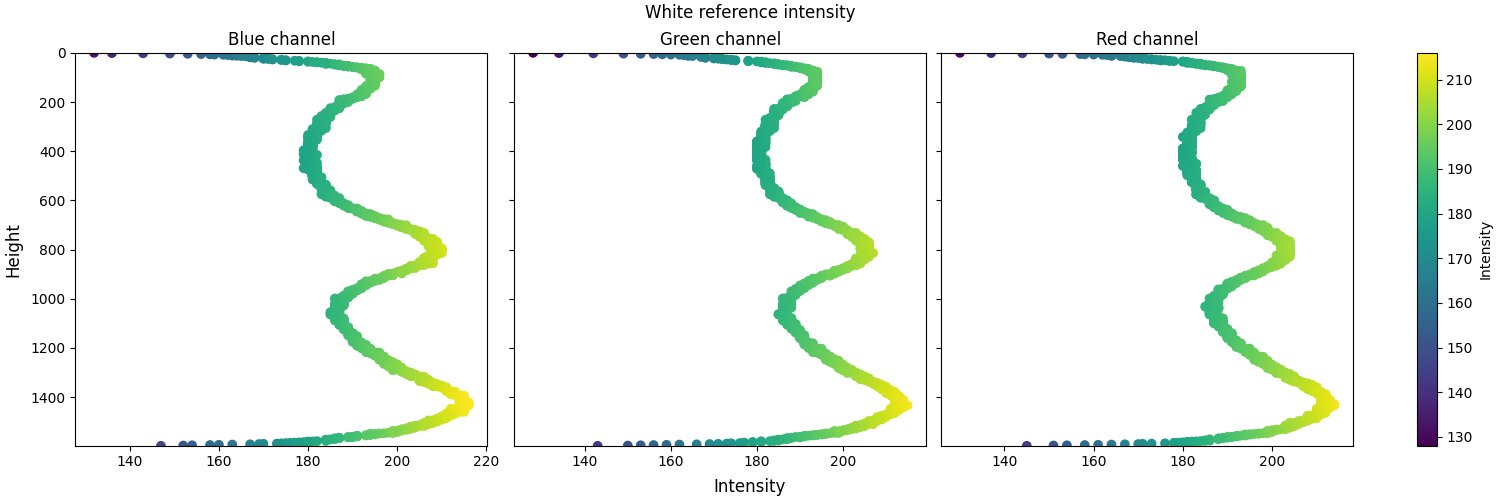}
        \caption{The row-wise third quantile of intensity of the white references from \myfigref{fig:reference_white_reference}. Any pixel in \myfigref{fig:reference_white_reference} that is not part of any white reference, as determined by \myfigref{fig:reference_raw_largest_components}, is ignored in this computation.}
        \label{fig:reference_white_reference_plot}
    \end{subfigure}
    
    \vspace{1em}
    
    % Second row
    \begin{subfigure}[t]{0.48\textwidth}
        \centering
        \includegraphics[width=\textwidth]{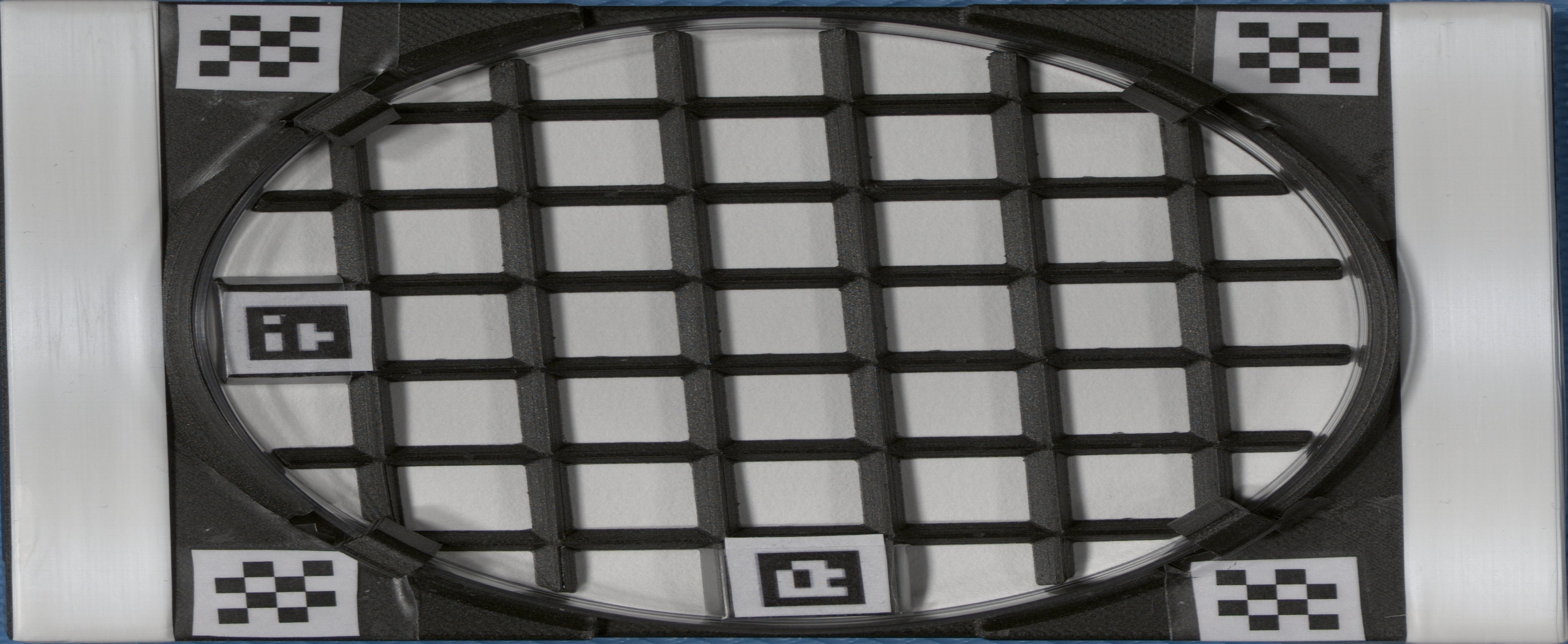}
        \caption{\myfigref{fig:raw_rgb_reference} cut off at the borders of the white references as determined by \myfigref{fig:reference_raw_largest_components}.}
        \label{fig:reference_plate}
    \end{subfigure}
    \hfill
    \begin{subfigure}[t]{0.48\textwidth}
        \centering
        \includegraphics[width=\textwidth]{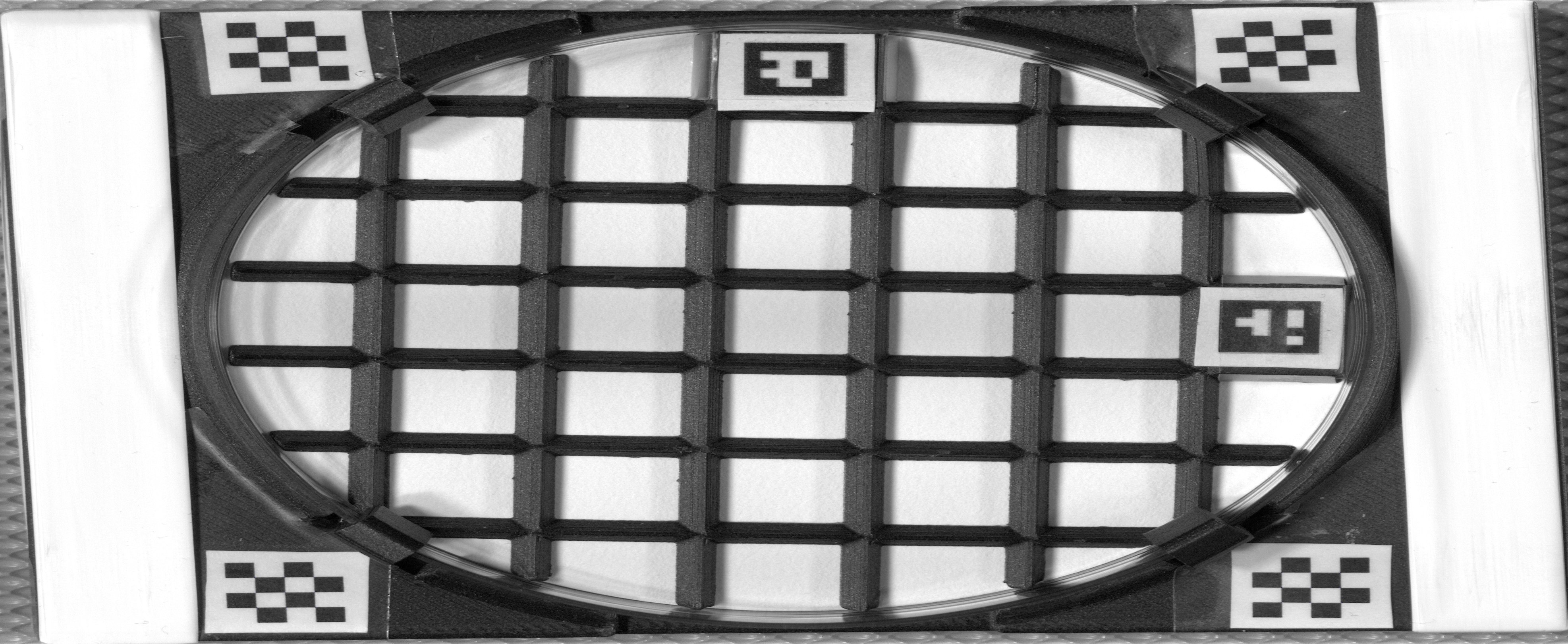}
        \caption{The image resulting from white-correction of \myfigref{fig:reference_plate} by \myeqnref{eq:white_dark_correction} using the values from \myfigref{fig:reference_white_reference} as shown in \myfigref{fig:reference_white_reference_plot}. This image is grayscaled to enable detection of the chessboards.}
        \label{fig:reference_plate_color_corrected}
    \end{subfigure}
\end{figure*}

\begin{figure*}[htbp]\ContinuedFloat
    \centering
    \begin{subfigure}[t]{0.4\textwidth}
        \centering
        \includegraphics[width=\textwidth]{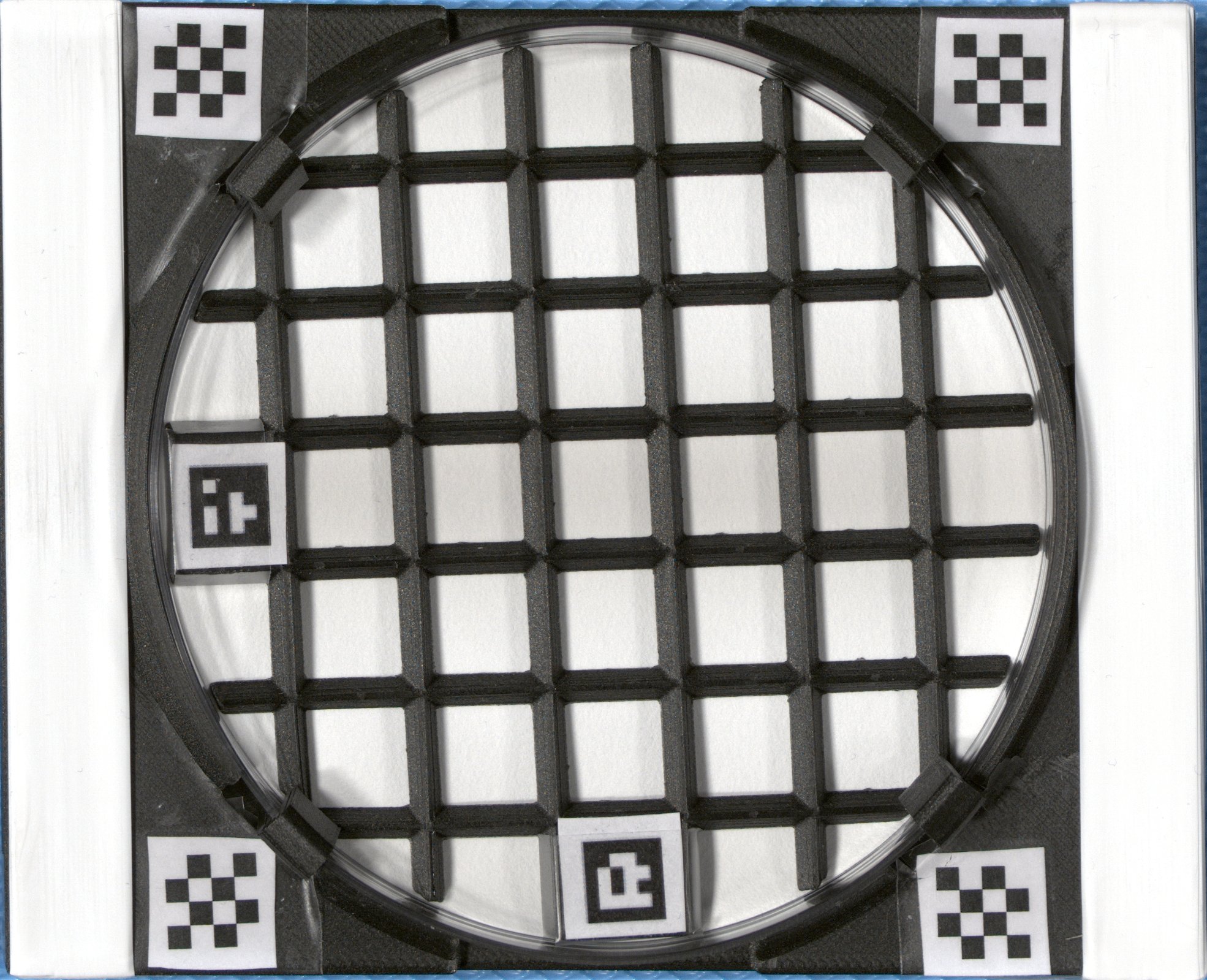}
        \caption{The image resulting from size-correction of the colorized version of \myfigref{fig:reference_plate_color_corrected} using bilinear interpolation with a factor based on the ratio between the height and width of the chessboard squares.}
        \label{fig:reference_plate_size_and_color_corrected}
    \end{subfigure}
    \hfill
    \begin{subfigure}[t]{0.4\textwidth}
        \centering
        \includegraphics[width=\textwidth]{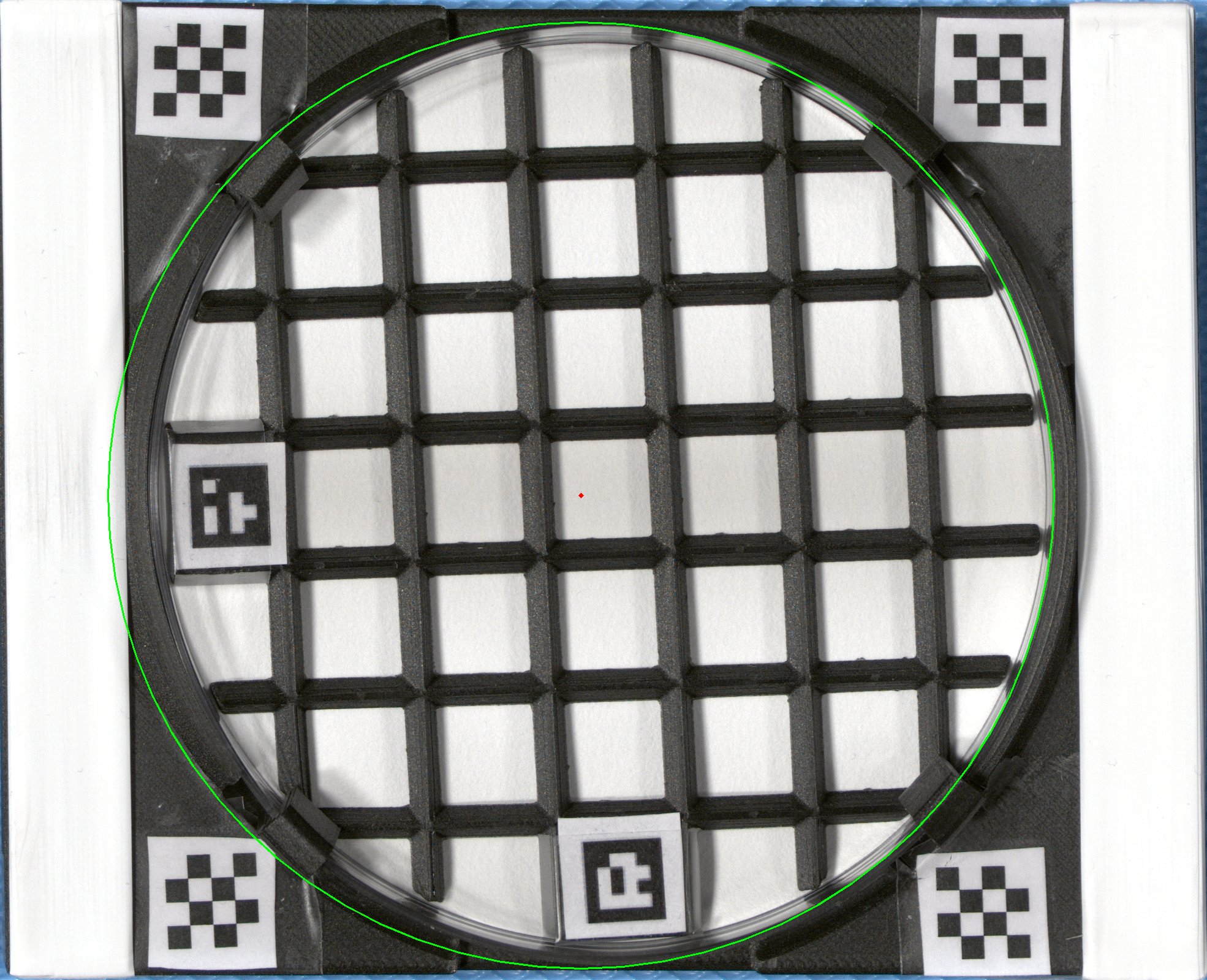}
        \caption{Approximating the location of the Petri dish in \myfigref{fig:reference_plate_size_and_color_corrected} with a Circle Hough Transform (\citealt{duda1972use}). The circle is shown with its border (green) and its center (red). Smoothing and edge detection (neither shown) are performed prior the Circle Hough Transform.}
        \label{fig:reference_hough_circle}
    \end{subfigure}

    \vspace{1em}

    \begin{subfigure}[t]{0.4\textwidth}
        \centering
        \includegraphics[width=\textwidth]{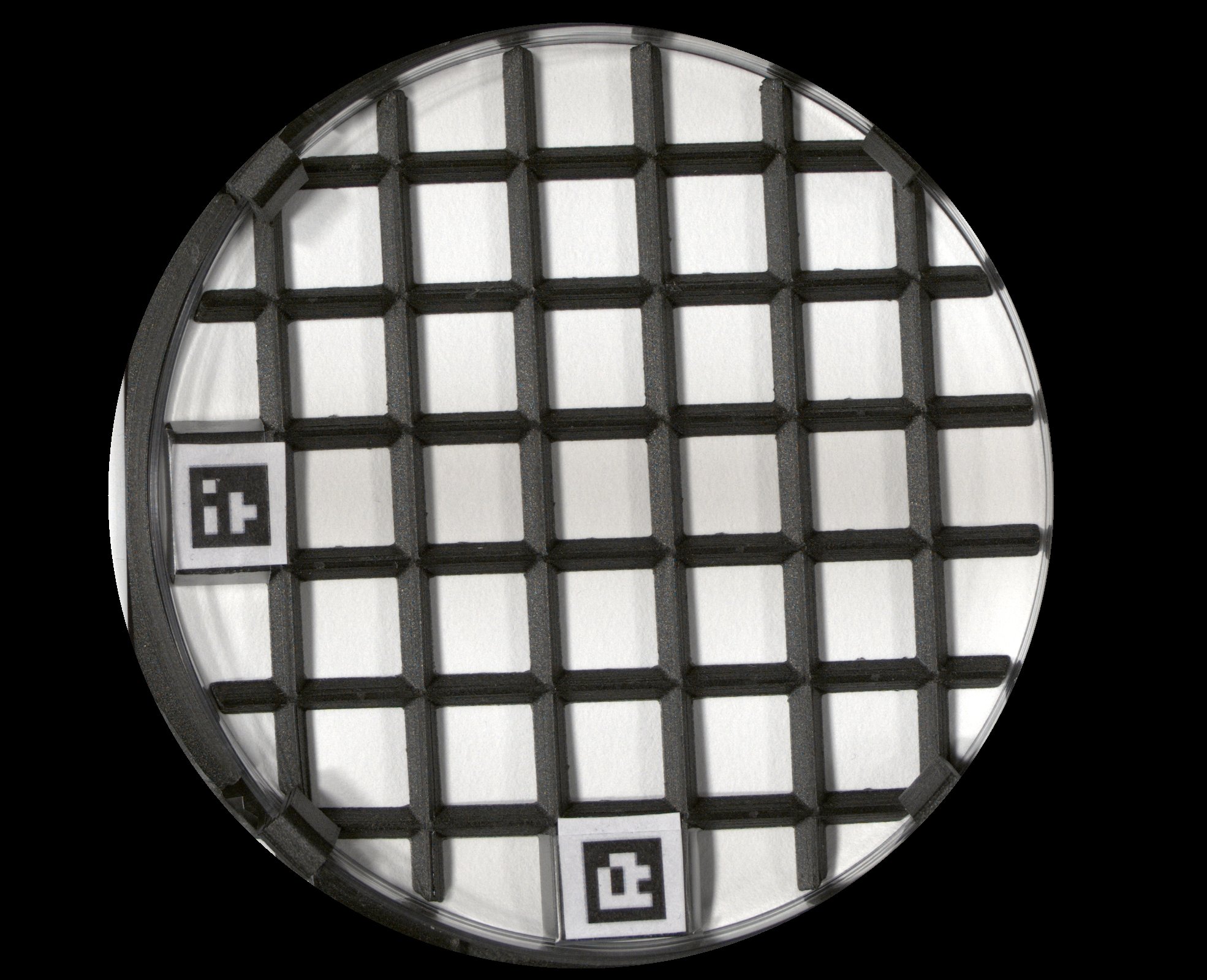}
        \caption{The image resulting from masking \myfigref{fig:reference_plate_size_and_color_corrected} with the circle found in \myfigref{fig:reference_hough_circle}.}
        \label{fig:reference_circle_mask}
    \end{subfigure}
    \hfill
    \centering
    \begin{subfigure}[t]{0.4\textwidth}
        \centering
        \includegraphics[width=\textwidth]{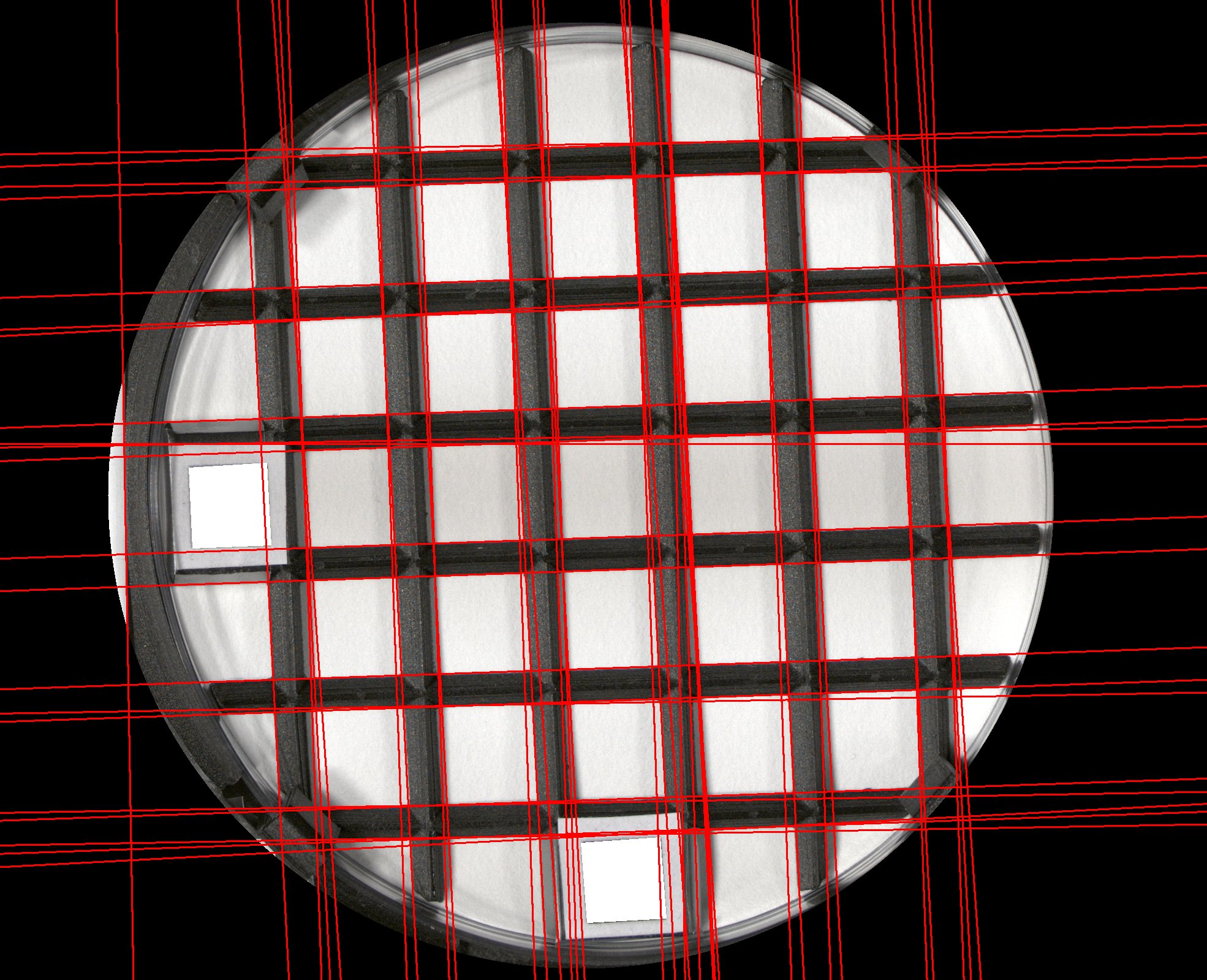}
        \caption{Masking the ArUco codes and performing a Hough Line Transform. Smoothing and edge detection (neither shown) are performed prior to the Hough Line Transform.}
        \label{fig:reference_hough_lines}
    \end{subfigure}

    \vspace{1em}
\end{figure*}

\begin{figure*}\ContinuedFloat
    \centering
    \begin{subfigure}[t]{0.4\textwidth}
        \centering
        \includegraphics[width=\textwidth]{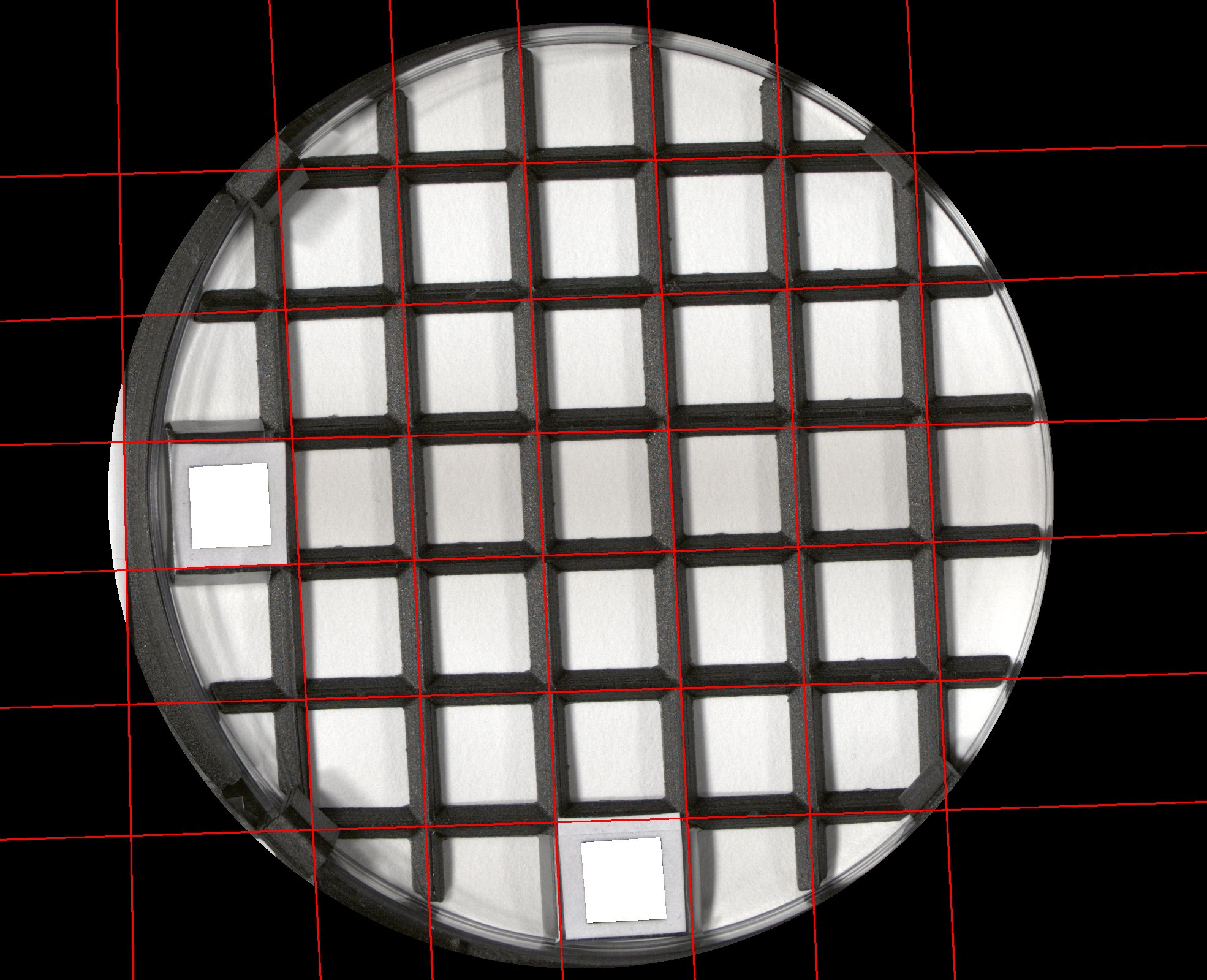}
        \caption{The result of averaging similar Hough lines from \myfigref{fig:reference_hough_lines}.}
        \label{fig:reference_avg_hough_lines}
    \end{subfigure}
    \hfill
    \begin{subfigure}[t]{0.4\textwidth}
        \centering
        \includegraphics[width=\textwidth]{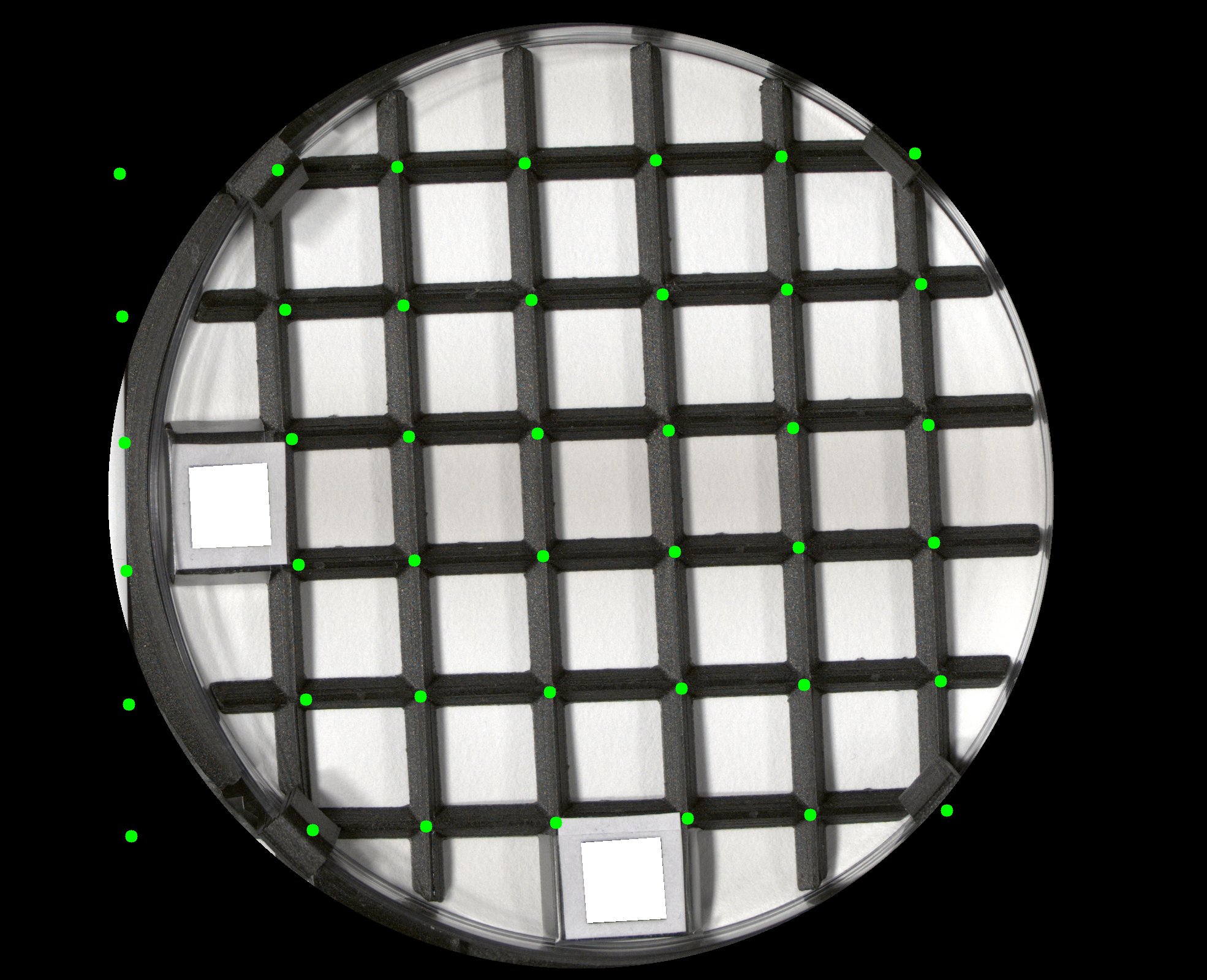}
        \caption{Find the intersections between all the lines in \myfigref{fig:reference_avg_hough_lines}. Discard any intersections that lie outside the image.}
        \label{fig:reference_intersections}
    \end{subfigure}
    \vspace{1em}
    \begin{subfigure}[t]{0.4\textwidth}
        \centering
        \includegraphics[width=\textwidth]{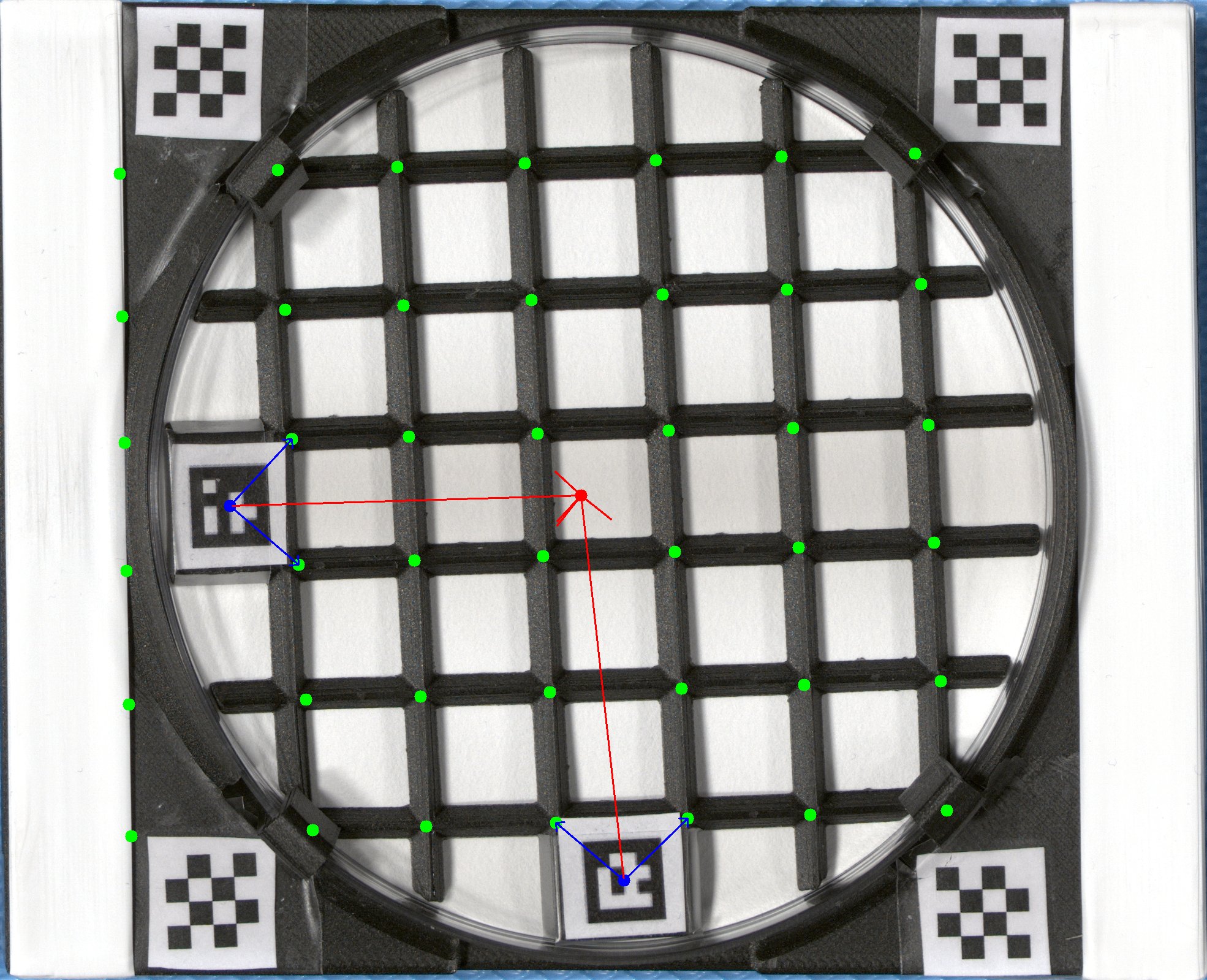}
        \caption{From the center of each ArUco marker, find the vector (shown in red) to the center of the circle found in \myfigref{fig:reference_hough_circle}. Then, for each ArUco code's center, find the two closest points that form an angle of less than 90 degrees with the corresponding red vectors (shown by blue vectors).}
        \label{fig:reference_vecs_img}
    \end{subfigure}
    \hfill
    \begin{subfigure}[t]{0.4\textwidth}
        \centering
        \includegraphics[width=\textwidth]{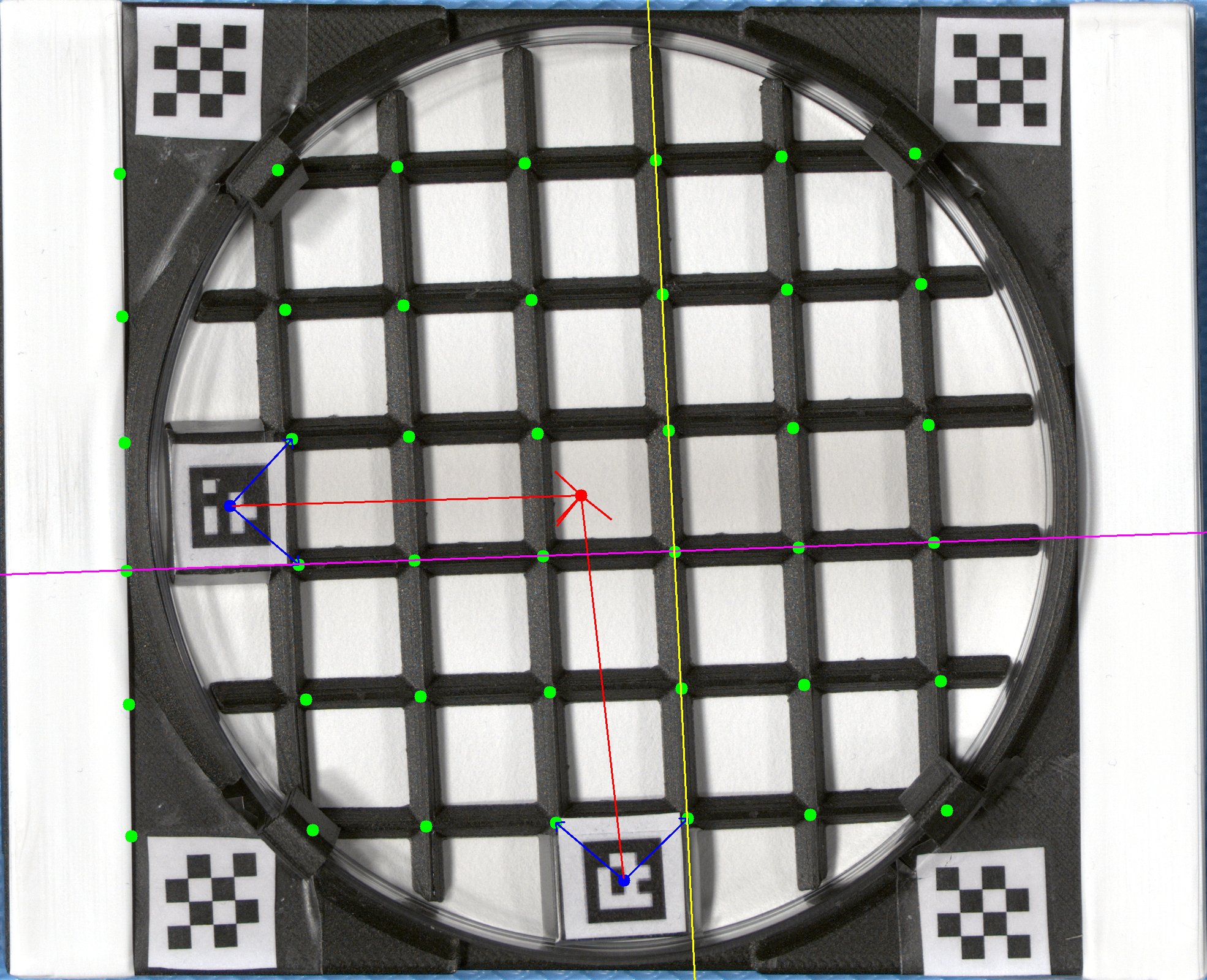}
        \caption{Each of the two pairs of points found in \myfigref{fig:reference_vecs_img} lie on a line found in \myfigref{fig:reference_avg_hough_lines}. Choose an arbitrary point from each of the two pairs and choose the line \emph{not shared by the other point from the same pair}. The magenta (yellow) line goes along the point chosen from the pair closest to the ArUco code with the lowest (highest) id.}
        \label{fig:reference_vecs_img_with_lines}
    \end{subfigure}
\end{figure*}

\begin{figure*}\ContinuedFloat
    \centering
    \begin{subfigure}[t]{0.4\textwidth}
        \centering
        \includegraphics[width=\textwidth]{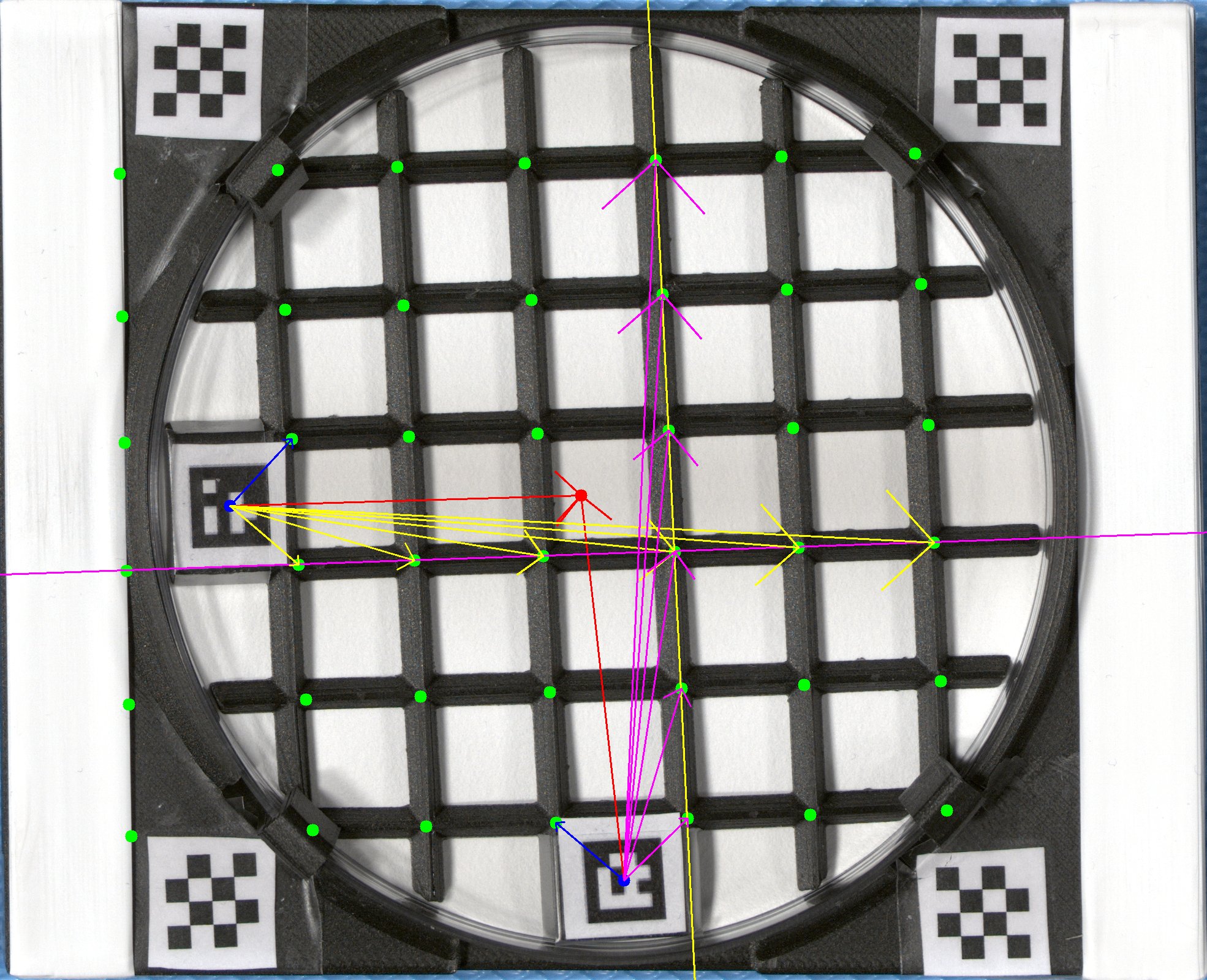}
        \caption{Find the intersections from \myfigref{fig:reference_intersections} that lie on the magenta (yellow) line found in \myfigref{fig:reference_vecs_img_with_lines}. Compute the vector from the center of the ArUco code with the highest (lowest) id to the intersections and discard any where the angle between the vector and the corresponding red vector (found in \myfigref{fig:reference_vecs_img}) is more than 90 degrees. The retained vectors are shown in yellow (magenta).}
        \label{fig:vecs_img_with_lines_and_intersections}
    \end{subfigure}
    \hfill
    \begin{subfigure}[t]{0.4\textwidth}
        \centering
        \includegraphics[width=\textwidth]{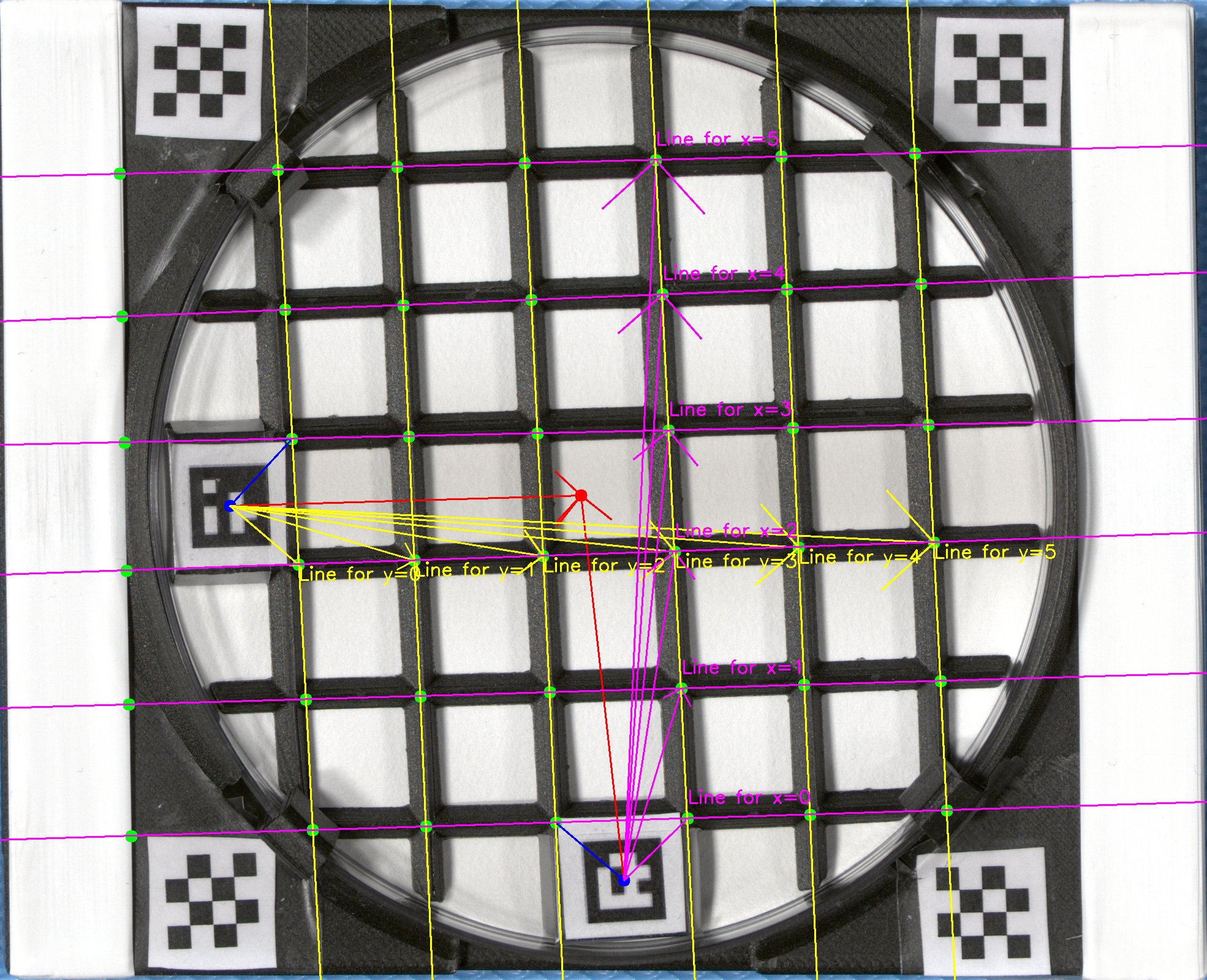}
        \caption{Enumerate the retained yellow (magenta) vectors from \myfigref{fig:vecs_img_with_lines_and_intersections} by their lengths. Each yellow (magenta) vector is the intersection between the magenta (yellow) line and a yellow (magenta) line. This intersecting yellow (magenta) line is where $y \in 
        \{0 \dots 5\}$ ($x \in \{0 \ldots 5\}$) as determined by its enumeration.}
        \label{fig:reference_grid_lines_img}
    \end{subfigure}
    \vspace{1em}
    \begin{subfigure}[t]{0.4\textwidth}
        \centering
        \includegraphics[width=\textwidth]{figures/pipeline/reference_img_26/grid_img.jpg}
        \caption{Identify the intersection points between the yellow and magenta lines from \myfigref{fig:reference_grid_lines_img} by their ($x,y$) coordinates. Then, adjust the location of each point to lie in a local minimum in intensity space (after grayscaling and smoothing, neither shown). The points in the corners and adjacent to the ArUco codes are not adjusted.}
        \label{fig:reference_grid_img}
    \end{subfigure}

    \caption{An overview of the image processing pipeline using Petri dish 26 as an example. Notice how the grid coordinates are uniquely defined by the detection in the RGB image in \myfigref{fig:reference_grid_rgb}. Localization of the same grid in a subsequent RGB or HSI image can be derived based on corresponding affine transformation matrices based on the ArUco codes' corners and the chessboards' centers, respectively.}
    \label{fig:reference_pipeline}
\end{figure*}

\subsection{RGB grid localization}\label{sec:rgb_grid_localization}

In this section, we explain how to use the result from \mysecref{sec:initial_grid_detection} to derive the location of the grid of Petri dish 26 in a subsequent RGB image of that same Petri dish. Here, we use the image acquired on day 1 as an example. The entire process is outlined in \myfigref{fig:pipeline_rgb_day_1}. As previously, we extract the plate from the image, perform white-correction with the white references, and use the chessboards to correct size. This takes us from the raw RGB camera output in \myfigref{fig:rgb_raw_rgb} to the white- and size-corrected plate image in \myfigref{fig:rgb_plate_size_and_color_corrected}. We then detect the ArUco codes' 8 corners on \myfigref{fig:rgb_plate_size_and_color_corrected} and use the robust RANSAC algorithm (\citealt{fischler1981ransac}) to estimate an affine transformation between those and the corresponding ArUco code corners in \myfigref{fig:reference_plate_size_and_color_corrected}. Given this transformation, we can directly transform the grid from \myfigref{fig:reference_grid_img} to our current RGB image (\myfigref{fig:pipeline_rgb_grid_day_1}). We can then use the coordinates to extract single grid cells and thus single barley kernels (e.g., \myfigref{fig:pipeline_rgb_single_kernel}).

\begin{figure*}
    \centering

    \begin{subfigure}[b]{\textwidth}
        \centering
        \includegraphics[width=\textwidth]{figures/pipeline/dish_26_rgb_day_1/raw_bgr_img.jpg}
        \caption{The raw image output by the RGB camera with the black filter paper and barley kernels.}
        \label{fig:rgb_raw_rgb}
    \end{subfigure}

    \vspace{1em}

    \begin{subfigure}[b]{\textwidth}
        \centering
        \includegraphics[width=\textwidth]{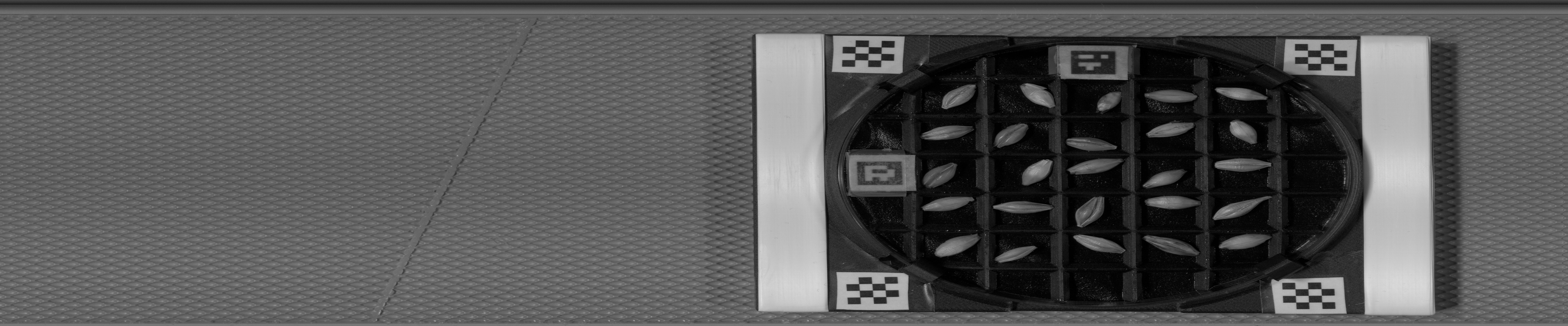}
        \caption{Grayscale version of \myfigref{fig:rgb_raw_rgb} achieved by averaging the color channels.}
        \label{fig:rgb_raw_gray}
    \end{subfigure}

    \vspace{1em}

    \begin{subfigure}[b]{\textwidth}
        \centering
        \includegraphics[width=\textwidth]{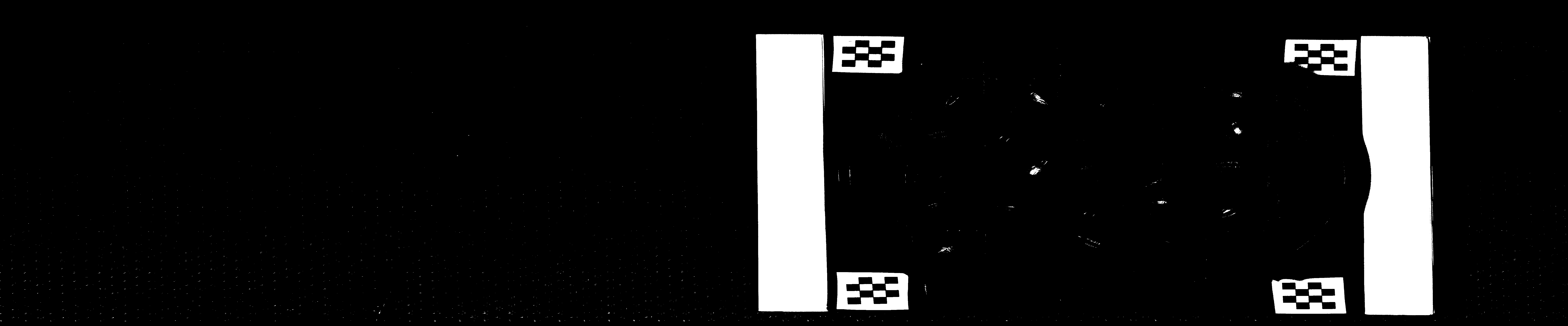}
        \caption{Binary version of \myfigref{fig:rgb_raw_gray} computed using Otsu's method (\citealt{Otsu79a}).}
        \label{fig:rgb_raw_binary}
    \end{subfigure}

\end{figure*}

\begin{figure*}\ContinuedFloat
    \centering

    \begin{subfigure}[b]{\textwidth}
        \centering
        \includegraphics[width=\textwidth]{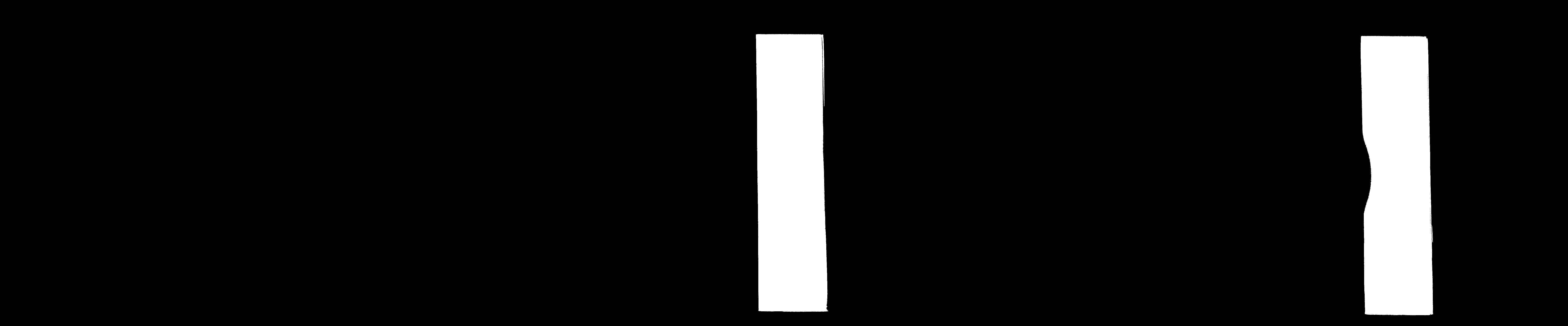}
        \caption{The two largest connected components from \myfigref{fig:rgb_raw_binary}. They correspond to the location of the two white references.}
        \label{fig:rgb_raw_largest_components}
    \end{subfigure}

    \vspace{1em}
    
    \begin{subfigure}[b]{\textwidth}
        \centering
        \includegraphics[width=\textwidth]{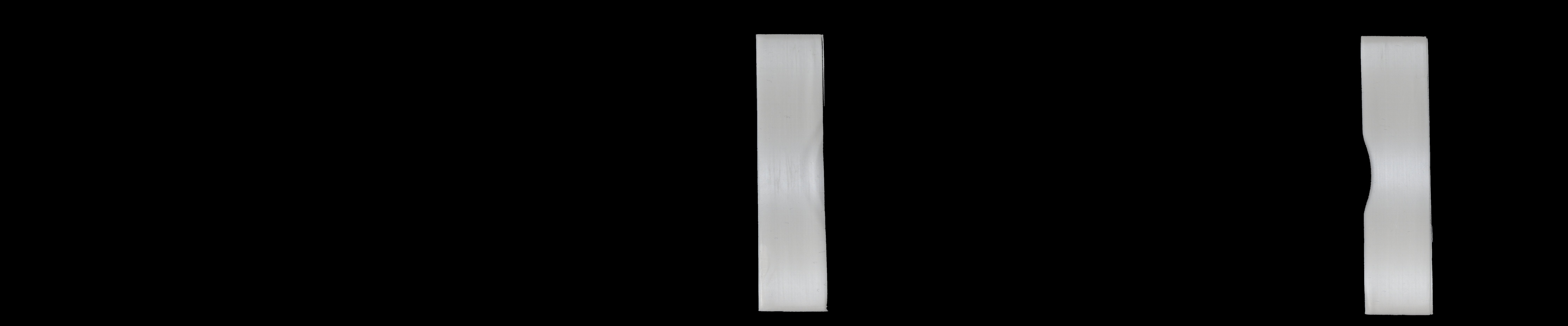}
        \caption{Applying \myfigref{fig:rgb_raw_largest_components} as a binary mask to \myfigref{fig:rgb_raw_rgb} to extract the white references.}
        \label{fig:rgb_white_reference}
    \end{subfigure}

    \vspace{1em}

    \begin{subfigure}[b]{\textwidth}
        \centering
        \includegraphics[width=\textwidth]{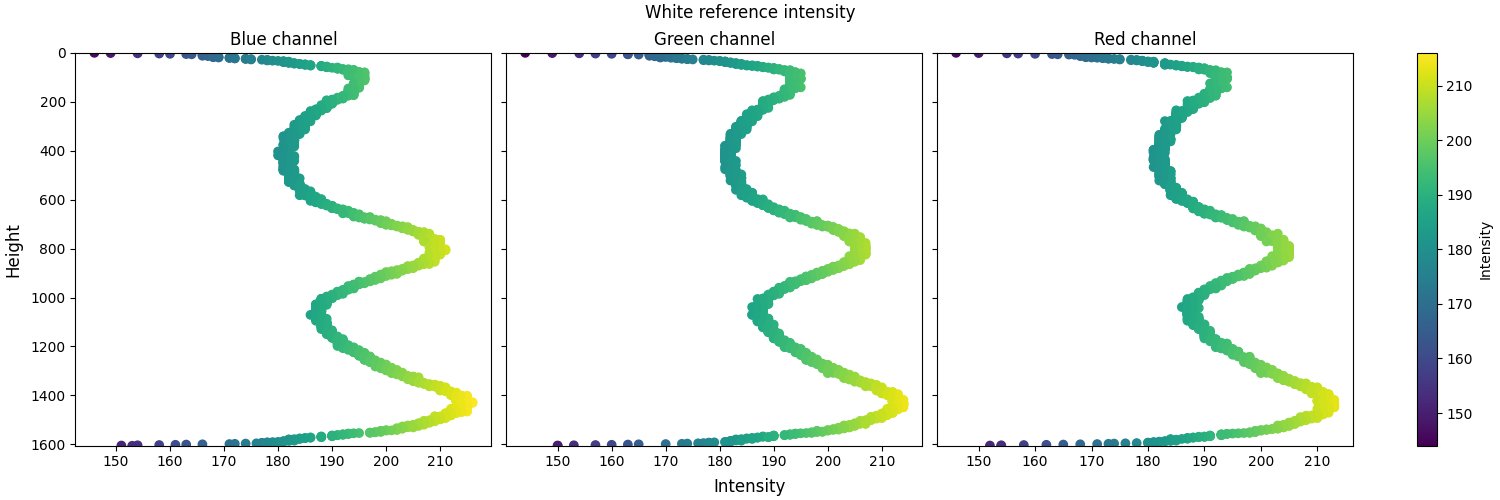}
        \caption{The row-wise third quantile of intensity of the white references from \myfigref{fig:rgb_white_reference}. Any pixel in \myfigref{fig:rgb_white_reference} that is not part of any white reference, as determined by \myfigref{fig:rgb_raw_largest_components}, is ignored in this computation. The 8-bit color channels give an intensity range between 0 and 255, both ends inclusive.}
        \label{fig:rgb_white_reference_plot}
    \end{subfigure}

    \vspace{1em}

    \begin{subfigure}[t]{0.48\textwidth}
        \centering
        \includegraphics[width=\textwidth]{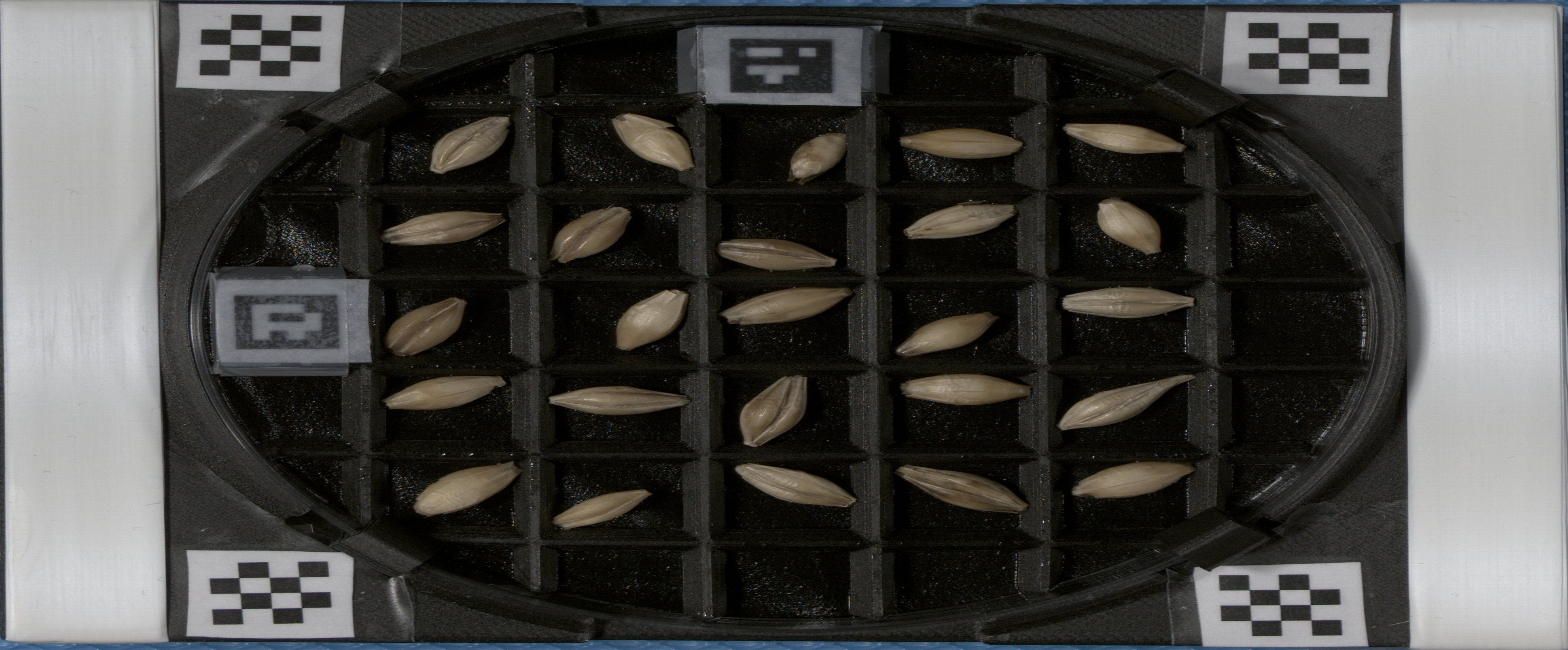}
        \caption{\myfigref{fig:rgb_raw_rgb} cut off at the borders of the white references as determined by \myfigref{fig:rgb_raw_largest_components}.}
        \label{fig:rgb_plate}
    \end{subfigure}
    \hfill
    \begin{subfigure}[t]{0.48\textwidth}
        \centering
        \includegraphics[width=\textwidth]{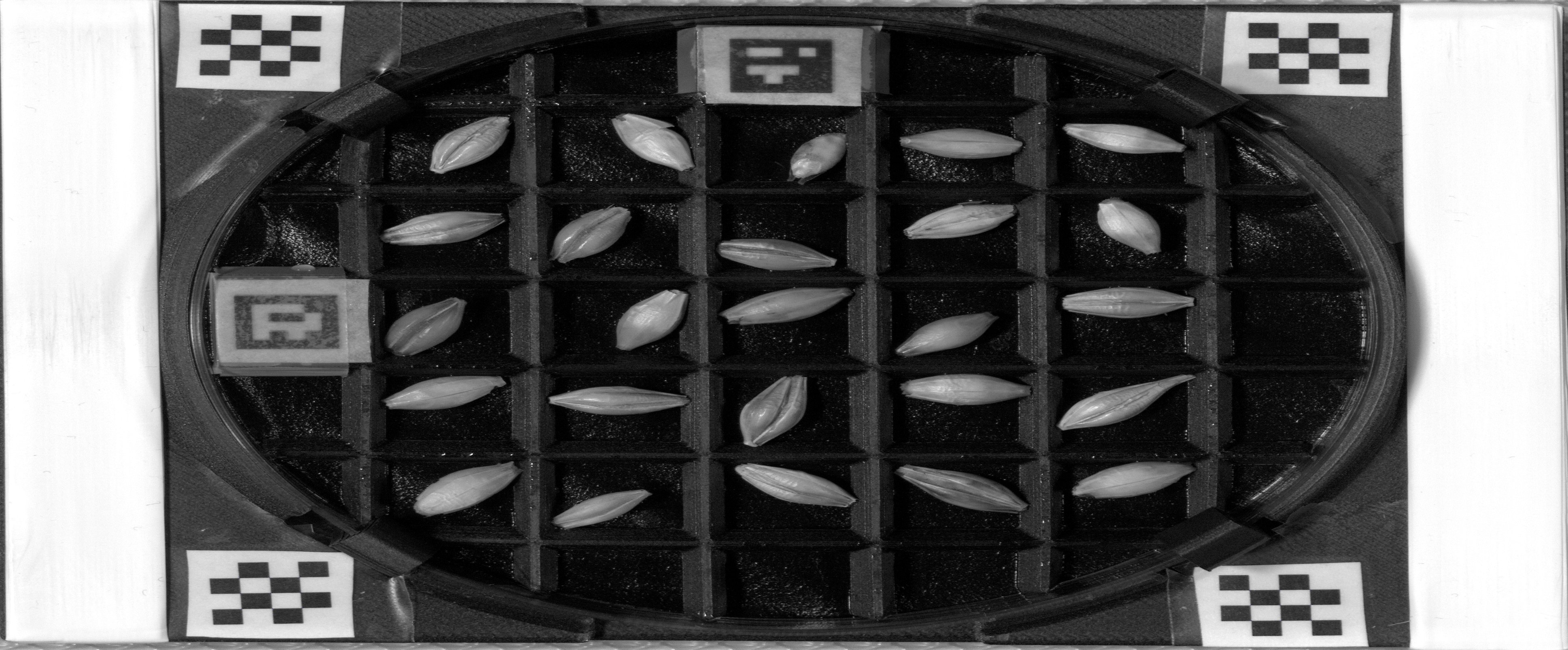}
        \caption{The image resulting from white-correction of \myfigref{fig:rgb_plate} by \myeqnref{eq:white_dark_correction} using the values from \myfigref{fig:rgb_white_reference} as shown in \myfigref{fig:rgb_white_reference_plot}. This image is grayscaled to enable detection of the chessboards.}
        \label{fig:rgb_plate_color_corrected}
    \end{subfigure}
    
\end{figure*}

\begin{figure*}\ContinuedFloat
    \centering

    \begin{subfigure}[t]{0.48\textwidth}
        \centering
        \includegraphics[width=\textwidth]{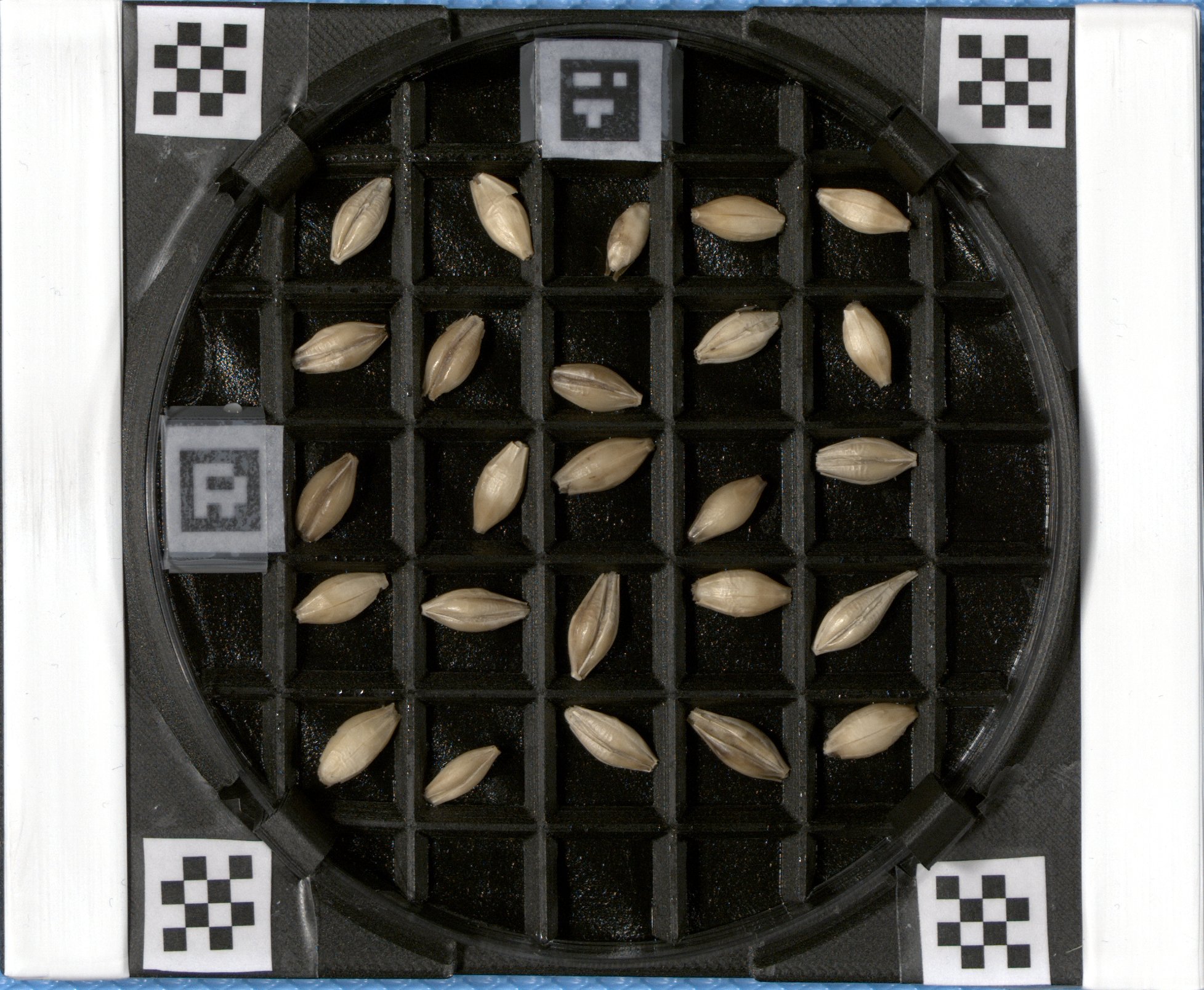}
        \caption{The image resulting from size-correction of the colorized version of \myfigref{fig:rgb_plate_color_corrected} using bilinear interpolation with a factor based on the ratio between the height and width of the chessboard squares.}
        \label{fig:rgb_plate_size_and_color_corrected}
    \end{subfigure}
    \hfill
    \begin{subfigure}[t]{0.48\textwidth}
        \centering
        \includegraphics[width=\textwidth]{figures/pipeline/dish_26_rgb_day_1/grid_img.jpg}
        \caption{The Petri dish from \myfigref{fig:rgb_raw_rgb}. Using the RANSAC algorithm (\citealt{fischler1981ransac}), an affine transformation matrix from \myfigref{fig:reference_grid_img} to \myfigref{fig:rgb_plate_size_and_color_corrected} is computed based on the 8 corners of the ArUco codes. This affine transformation matrix is then used to move the grid from \myfigref{fig:reference_grid_img} to \myfigref{fig:rgb_plate_size_and_color_corrected} without having to redetect the grid.}
        \label{fig:pipeline_rgb_grid_day_1}
    \end{subfigure}

    \vspace{1em}
    
    \begin{subfigure}[b]{0.48\textwidth}  % First image aligned with the left column
        \centering
        \includegraphics[width=0.1\textwidth]{figures/pipeline/dish_26_rgb_day_1/grain_0.jpg}  % Smaller image
        \caption{A cutout of the grid cell in \myfigref{fig:pipeline_rgb_grid_day_1} defined by a polygon between the grid points $\{(0,0), (1,0), (1,1), (0,1)\}$.}
        \label{fig:pipeline_rgb_single_kernel}
    \end{subfigure}

    \caption{An overview of the image processing pipeline using Petri dish 26 as an example. This image is taken 1 day after exposure to moisture. The stage of the pipeline shown here is that of localization of the grid shown in \myfigref{fig:pipeline_rgb_grid_day_1}. Due to the previous detection of the same grid in \myfigref{fig:reference_grid_img}, localization of the grid in this image only requires detection of the ArUco codes.}
    \label{fig:pipeline_rgb_day_1}
\end{figure*}

\subsection{HSI grid localization}\label{sec:hsi_grid_localization}

In this section, we explain how to use the result from \mysecref{sec:rgb_grid_localization} to derive the location of the grid of Petri dish 26 in the corresponding HSI image of Petri dish 26 acquired on day 1. The entire process is outlined in \myfigref{fig:pipeline_hsi_day_1}. As with the RGB images, we use a combination of Otsu thresholding (\citealt{Otsu79a}) and connected components to derive the white references, which are then used to detect the plate and white correction. However, unlike the RGB camera, the HSI camera has a non-zero dark signal that must be corrected. To measure the dark signal, we take an image with the camera's shutter closed and compute the row-wise median of that dark image. Then, to ensure alignment with the white reference signal, we retain only the rows of the dark signal corresponding to the rows of the HSI image that also contain at least one pixel of white reference. The intensity (white and dark) correction is then computed based on \myeqnref{eq:white_dark_correction} using the values shown in \myfigref{fig:hsi_white_ref_cbar} and \myfigref{fig:hsi_dark_ref_cbar}.

Subsequent size correction using bilinear interpolation with a factor based on the size of the chessboard squares is then applied to standardize the pixel size. Although less profound than with the RGB images, the width of the chessboards is larger than their height. This, coupled with the fact that the HSI camera acquires one column of the image at a time (line scan), ensures no loss of information when downsizing the image. The result is shown in \myfigref{fig:hsi_plate_img_size_and_color_corrected}. Using the RANSAC algorithm (\citealt{fischler1981ransac}), we then use the centers of the four chessboards in \myfigref{fig:hsi_plate_img_size_and_color_corrected} and those in \myfigref{fig:rgb_plate_size_and_color_corrected} to transform the grid from \myfigref{fig:pipeline_rgb_grid_day_1} to \myfigref{fig:pipeline_hsi_grid_day_1}. This, in turn, allows us to cut out individual grid cells (and thus barley kernels) for subsequent analysis. For example, the same physical barley kernel is imaged at the same time in RGB (\myfigref{fig:pipeline_rgb_single_kernel}) and HSI (\myfigref{fig:pipeline_rgb_single_kernel}), enabling direct comparison between the two modalities.

\begin{figure*}
    \centering

    \begin{subfigure}[b]{\textwidth}
        \centering
        \includegraphics[width=\textwidth]{figures/pipeline/dish_26_hsi_day_1/raw_gray_img.jpg}
        \caption{Raw output from the HSI camera, converted to grayscale for visualization purposes.}
        \label{fig:pipeline_raw_hsi_gray}
    \end{subfigure}

    \vspace{1em}

    \begin{subfigure}[b]{\textwidth}
        \centering
        \includegraphics[width=\textwidth]{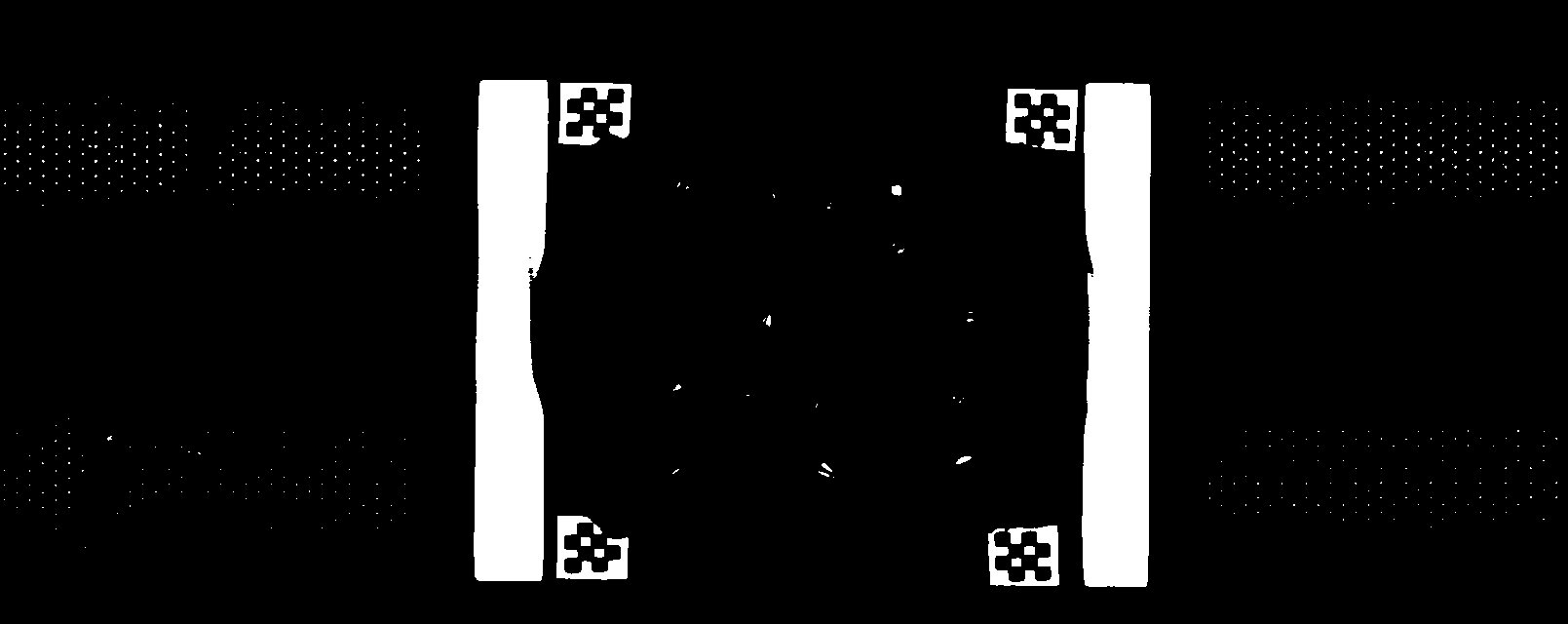}
        \caption{A binary version of \myfigref{fig:pipeline_raw_hsi_gray} computed using Otsu's method (\citealt{Otsu79a}).}
        \label{fig:pipeline_raw_hsi_binary}
    \end{subfigure}

    \vspace{1em}
    
    \begin{subfigure}[b]{\textwidth}
        \centering
        \includegraphics[width=\textwidth]{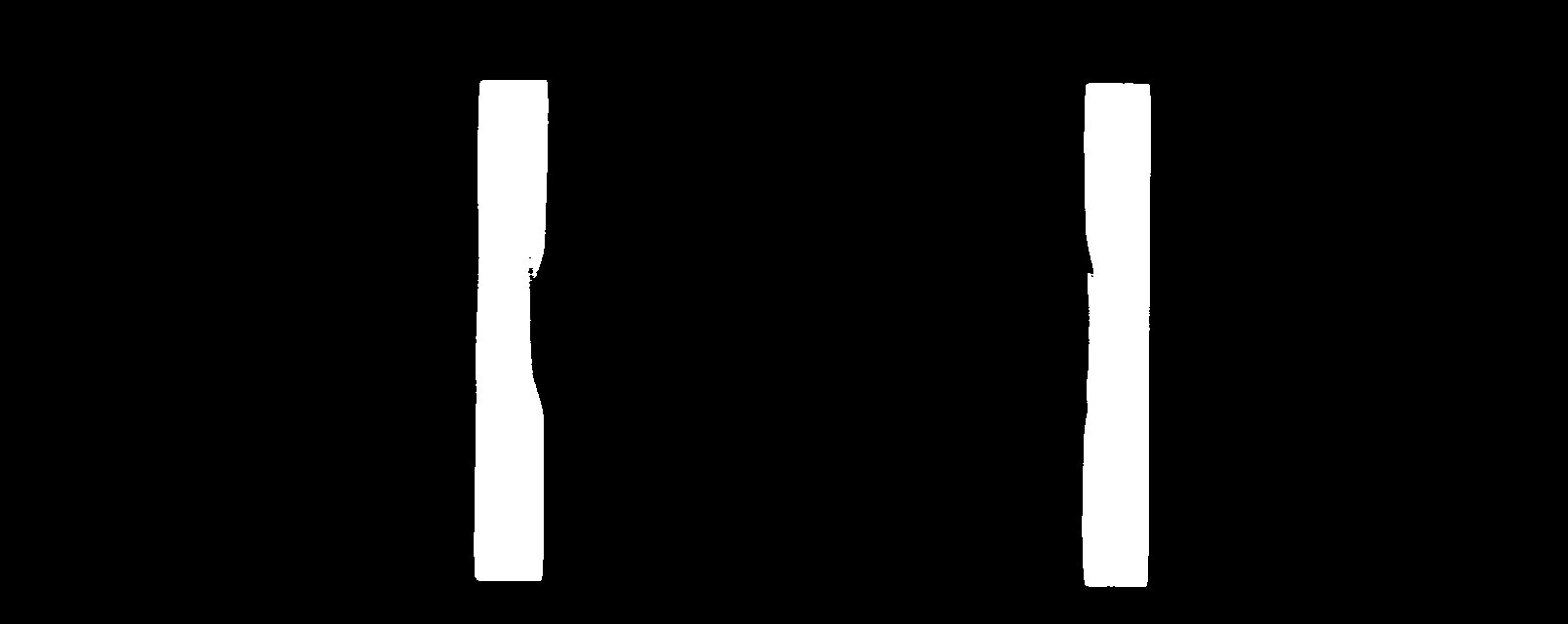}
        \caption{The two largest connected components from \myfigref{fig:pipeline_raw_hsi_binary}. They correspond to the location of the two white references.}
        \label{fig:hsi_raw_largest_components}
    \end{subfigure}
    
\end{figure*}

\begin{figure*}\ContinuedFloat
    \centering
    
    \begin{subfigure}[b]{\textwidth}
        \centering
        \includegraphics[width=\textwidth]{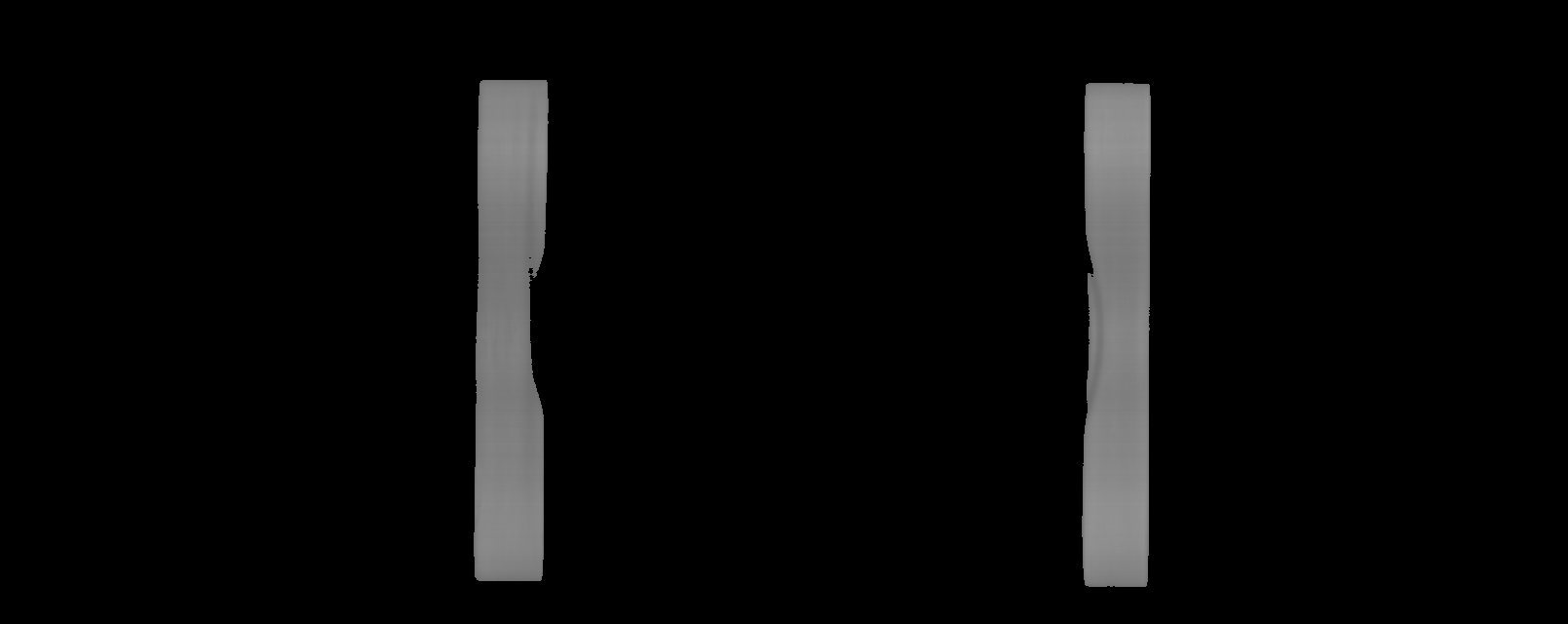}
        \caption{Applying \myfigref{fig:hsi_raw_largest_components} as a binary mask to the HSI version of \myfigref{fig:pipeline_raw_hsi_gray} to extract the white references.}
        \label{fig:hsi_raw_largest_components_overlap}
    \end{subfigure}

    \vspace{1em}

    \begin{subfigure}[t]{0.48\textwidth}
        \centering
        \includegraphics[width=\textwidth]{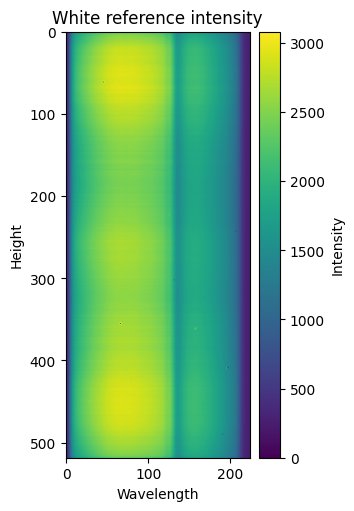}
        \caption{The row-wise third quantile of intensity of the white references from \myfigref{fig:hsi_raw_largest_components_overlap}. Any pixel in \myfigref{fig:hsi_raw_largest_components_overlap} that is not part of any white reference, as determined by \myfigref{fig:hsi_raw_largest_components}, is ignored in this computation. The 12-bit spectral channels give an intensity range between 0 and 4095, both ends inclusive.}
        \label{fig:hsi_white_ref_cbar}
    \end{subfigure}
    \hfill
    \begin{subfigure}[t]{0.48\textwidth}
        \centering
        \includegraphics[width=\textwidth]{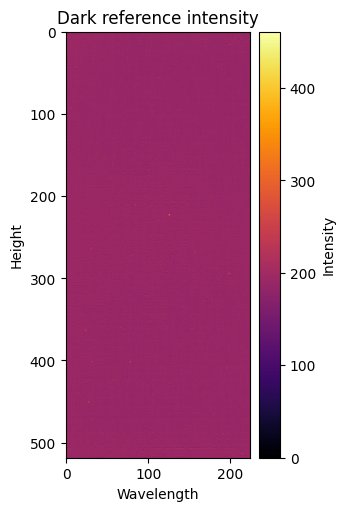}
        \caption{The row-wise median of intensity of a dark reference image taken with the HSI cameras' shutter closed. We retain only rows of the dark image corresponding to the rows where the white reference is located as determined by \myfigref{fig:hsi_raw_largest_components}. This ensures consistency between the white and dark correction. The 12-bit spectral channels give an intensity range between 0 and 4095, both ends inclusive.}
        \label{fig:hsi_dark_ref_cbar}
    \end{subfigure}
    
\end{figure*}

\begin{figure*}\ContinuedFloat
    \centering

    \begin{subfigure}[t]{0.48\textwidth}
        \centering
        \includegraphics[width=\textwidth]{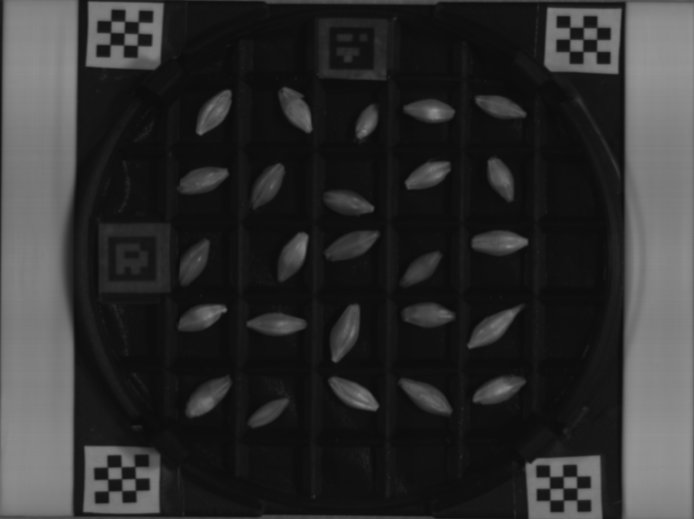}
        \caption{\myfigref{fig:pipeline_raw_hsi_gray} cut off at the borders of the white references as determined by \myfigref{fig:rgb_raw_largest_components}.}
        \label{fig:hsi_plate_img}
    \end{subfigure}
    \hfill
    \begin{subfigure}[t]{0.48\textwidth}
        \centering
        \includegraphics[width=\textwidth]{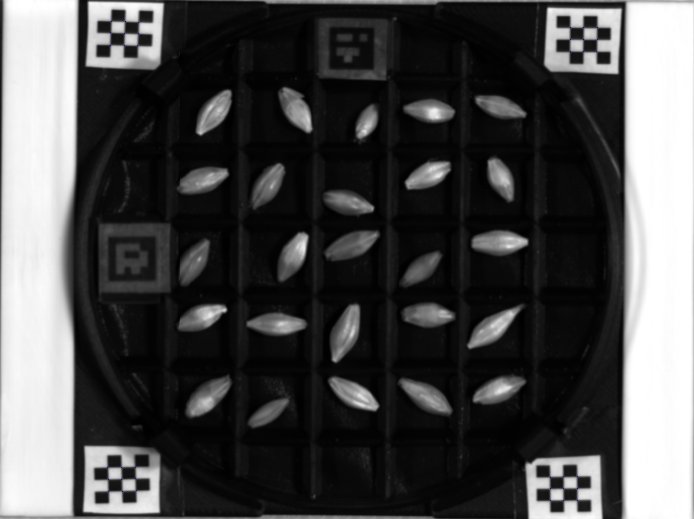}
        \caption{The image resulting from white- and dark-correction of the HSI version of \myfigref{fig:pipeline_raw_hsi_gray} using the values from \myfigref{fig:hsi_white_ref_cbar} and \myfigref{fig:hsi_dark_ref_cbar}. This image is grayscaled to enable detection of the chessboards.}
        \label{fig:hsi_plate_color_corrected}
    \end{subfigure}

    \vspace{1em}

    \begin{subfigure}[t]{0.48\textwidth}
        \centering
        \includegraphics[width=\textwidth]{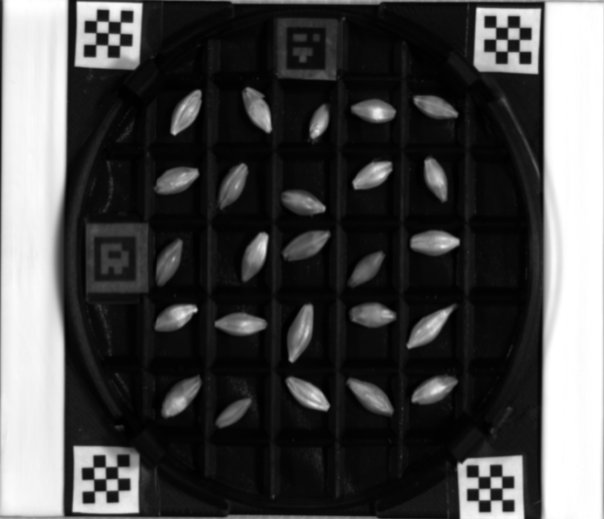}
        \caption{Grayscale version of the image resulting from size-correction of the HSI version of \myfigref{fig:hsi_plate_color_corrected} using bilinear interpolation with a factor based on the chessboard squares.}
        \label{fig:hsi_plate_img_size_and_color_corrected}
    \end{subfigure}
    \hfill
    \begin{subfigure}[t]{0.48\textwidth}
        \centering
        \includegraphics[width=\textwidth]{figures/pipeline/dish_26_hsi_day_1/grid_img.jpg}
        \caption{The Petri dish from \myfigref{fig:pipeline_raw_hsi_gray}. Using the RANSAC algorithm (\citealt{fischler1981ransac}), an affine transformation matrix from \myfigref{fig:pipeline_rgb_grid_day_1} to \myfigref{fig:hsi_plate_img_size_and_color_corrected} is computed based on the four chessboard centers. This affine transformation matrix is then used to move the grid from \myfigref{fig:pipeline_rgb_grid_day_1} to \myfigref{fig:hsi_plate_img_size_and_color_corrected} without having to redetect the grid.}
        \label{fig:pipeline_hsi_grid_day_1}
    \end{subfigure}
\end{figure*}

\begin{figure*}
    \vspace{1em}
    
    \begin{subfigure}[b]{0.48\textwidth}  % First image aligned with the left column
        \centering
        \includegraphics[width=0.1\textwidth]{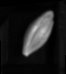}  % Smaller image
        \caption{A grayscaled cutout of the grid cell in \myfigref{fig:pipeline_hsi_grid_day_1} defined by a polygon between the grid points $\{(0,0), (1,0), (1,1), (0,1)\}$.}
        \label{fig:pipeline_hsi_single_kernel}
    \end{subfigure}

    \caption{An overview of the image processing pipeline using Petri dish 26 as an example. This image is taken 1 day after exposure to moisture. The stage of the pipeline shown here is that of localization of the grid shown in \myfigref{fig:pipeline_hsi_grid_day_1}. Due to the previous detection of the same grid in \myfigref{fig:reference_grid_img} and subsquently in \myfigref{fig:pipeline_rgb_grid_day_1}, localization of the grid in this image only requires detection of the chessboards.}
    \label{fig:pipeline_hsi_day_1}
\end{figure*}

\end{document}